\documentclass[10pt,logo,copyright]{giga-report}
\usepackage[authoryear,sort&compress,round]{natbib}

\usepackage[utf8]{inputenc} %
\usepackage[T1]{fontenc}    %

\usepackage{parskip}        %
\usepackage{url}            %
\usepackage{booktabs}       %
\usepackage{amsfonts}       %
\usepackage{nicefrac}       %
\usepackage{microtype}      %
\usepackage{xcolor}         %
\usepackage[dvipsnames]{xcolor} %
\usepackage{graphicx}
\usepackage{animate}        %
\usepackage{subcaption}
\usepackage{tabularx}
\usepackage{makecell}
\usepackage{adjustbox}
\usepackage{setspace}
\usepackage{todonotes}
\usepackage{colortbl}
\usepackage{wrapfig}

\newcolumntype{M}[1]{>{\centering\arraybackslash}m{#1}}
\usepackage{float}
\usepackage{tikz}
\usetikzlibrary{positioning,shapes,arrows}
\usepackage{amsmath,amsfonts,bm, bbm,leftindex}
\usepackage{multirow}
\usepackage{comment}
\usepackage{gensymb}
\usepackage{lipsum}
\usetikzlibrary{arrows.meta, positioning, fit}
\usepackage[para]{threeparttable}
\usepackage{tikz}
\usetikzlibrary{tikzmark}

\def\eqref#1{equation~\ref{#1}}

\def\1{\bm{1}}

\DeclareMathAlphabet{\mathsfit}{\encodingdefault}{\sfdefault}{m}{sl}
\SetMathAlphabet{\mathsfit}{bold}{\encodingdefault}{\sfdefault}{bx}{n}

\makeatletter
\let\save@mathaccent\mathaccent
\newcommand*\if@single[3]{%
  \setbox0\hbox{${\mathaccent"0362{#1}}^H$}%
  \setbox2\hbox{${\mathaccent"0362{\kern0pt#1}}^H$}%
  \ifdim\ht0=\ht2 #3\else #2\fi
  }
\newcommand*\rel@kern[1]{\kern#1\dimexpr\macc@kerna}
\newcommand*\widebar[1]{\@ifnextchar^{{\wide@bar{#1}{0}}}{\wide@bar{#1}{1}}}
\newcommand*\wide@bar[2]{\if@single{#1}{\wide@bar@{#1}{#2}{1}}{\wide@bar@{#1}{#2}{2}}}
\newcommand*\wide@bar@[3]{%
  \begingroup
  \def\mathaccent##1##2{%
    \let\mathaccent\save@mathaccent
    \if#32 \let\macc@nucleus\first@char \fi
    \setbox\z@\hbox{$\macc@style{\macc@nucleus}_{}$}%
    \setbox\tw@\hbox{$\macc@style{\macc@nucleus}{}_{}$}%
    \dimen@\wd\tw@
    \advance\dimen@-\wd\z@
    \divide\dimen@ 3
    \@tempdima\wd\tw@
    \advance\@tempdima-\scriptspace
    \divide\@tempdima 10
    \advance\dimen@-\@tempdima
    \ifdim\dimen@>\z@ \dimen@0pt\fi
    \rel@kern{0.6}\kern-\dimen@
    \if#31
      \overline{\rel@kern{-0.6}\kern\dimen@\macc@nucleus\rel@kern{0.4}\kern\dimen@}%
      \advance\dimen@0.4\dimexpr\macc@kerna
      \let\final@kern#2%
      \ifdim\dimen@<\z@ \let\final@kern1\fi
      \if\final@kern1 \kern-\dimen@\fi
    \else
      \overline{\rel@kern{-0.6}\kern\dimen@#1}%
    \fi
  }%
  \macc@depth\@ne
  \let\math@bgroup\@empty \let\math@egroup\macc@set@skewchar
  \mathsurround\z@ \frozen@everymath{\mathgroup\macc@group\relax}%
  \macc@set@skewchar\relax
  \let\mathaccentV\macc@nested@a
  \if#31
    \macc@nested@a\relax111{#1}%
  \else
    \def\gobble@till@marker##1\endmarker{}%
    \futurelet\first@char\gobble@till@marker#1\endmarker
    \ifcat\noexpand\first@char A\else
      \def\first@char{}%
    \fi
    \macc@nested@a\relax111{\first@char}%
  \fi
  \endgroup
}
\makeatother

\definecolor{darkred}{rgb}{0.7, 0.0, 0.0}

\usepackage{makecell}
\usepackage{tabularx}
\usepackage{graphicx}
\usepackage{booktabs}
\usepackage{multirow}
\usepackage{wrapfig}
\usepackage{array}
\usepackage{enumitem}
\usepackage{booktabs}    
\usepackage{graphicx}    
\usepackage[table]{xcolor} 
\definecolor{mygray}{gray}{0.9} 

\newcommand{\best}[1]{\textbf{#1}}
\newcommand{\secondbest}[1]{\underline{#1}}

\usepackage{amsmath} 

\usepackage[table]{xcolor}

\usepackage{pifont}
\usepackage{graphicx}

\usepackage[nameinlink]{cleveref}
\crefname{equation}{Eq.}{Eqs.}
\crefname{figure}{Fig.}{Figs.}
\crefname{section}{Sec.}{Sec.}
\crefname{appendix}{App.}{App.}
\crefname{table}{Tab.}{Tabs.}
\crefname{algorithm}{Algo}{Algo}
\crefname{thm}{Thm}{Thm}
\Crefname{thm}{Thm}{Thm}
\crefname{prop}{Prop}{Prop}

\newcommand{\crefnames}[3]{%
  \@for\next:=#1\do{%
    \expandafter\crefname\expandafter{\next}{#2}{#3}%
  }%
}

\title{GigaWorld-Policy: An Efficient Action-Centered World--Action Model}

\author{
\vspace{-0.1in}
\centerline{GigaAI} 
\centerline{{Project Page: \href{https://gigaai-research.github.io/GigaWorld-Policy/}{https://gigaai-research.github.io/GigaWorld-Policy/}}} 
\footnotesize
\textbf{GigaWorld Team (alphabetical order)}:
\normalfont
Angen Ye,
Boyuan Wang,
Chaojun Ni,
Guan Huang,
Guosheng Zhao,
Hao Li,
Hengtao Li,
Jie Li,
Jindi Lv,
Jingyu Liu,
Min Cao,
Peng Li,
Qiuping Deng,
Wenjun Mei,
Xiaofeng Wang,
Xinze Chen,
Xinyu Zhou,
Yang Wang,
Yifan Chang,
Yifan Li,
Yukun Zhou,
Yun Ye,
Zhichao Liu,
Zheng Zhu.

\vspace{-1em}
}

\begin{document}
\maketitle


\begin{abstract}
World--Action Models~(WAM) initialized from pre-trained video generation backbones have demonstrated remarkable potential for robot policy learning. However, existing approaches face two critical bottlenecks that hinder performance and deployment. First, jointly reasoning over future visual dynamics and corresponding actions incurs substantial inference overhead. Second, joint modeling often entangles visual and motion representations, making motion prediction accuracy heavily dependent on the quality of future video forecasts. To address these issues, we introduce GigaWorld-Policy, an action-centered WAM that learns 2D pixel–action dynamics while enabling efficient action decoding, with optional video generation. Specifically, we formulate policy training into two coupled components: the model predicts future action sequences conditioned on the current observation, and simultaneously generates future videos conditioned on the predicted actions and the same observation. The policy is supervised by both action prediction and video generation, providing richer learning signals and encouraging physically plausible actions through visual-dynamics constraints. With a causal design that prevents future-video tokens from influencing action tokens, explicit future-video generation is optional at inference time, allowing faster action prediction during deployment. To support this paradigm, we curate a diverse, large-scale robot dataset to pre-train an action-centered video generation model, which is then adapted as the backbone for robot policy learning. Experimental results on real-world robotic platforms show that GigaWorld-Policy runs $9\times$ faster than the leading WAM baseline, Motus, while improving task success rates by $7\%$.
Moreover, compared with $\pi_{0.5}$, GigaWorld-Policy improves performance by $95\%$ on RoboTwin~2.0.

\keywords{Video Generation\and World--Action Models \and Robot Policy Learning \and Large-Scale Robot Dataset}

\end{abstract}

\section{INTRODUCTION}
\begin{wrapfigure}{r}{0.45\linewidth}
    \centering
     \vspace{-3mm}
    \includegraphics[width=\linewidth]{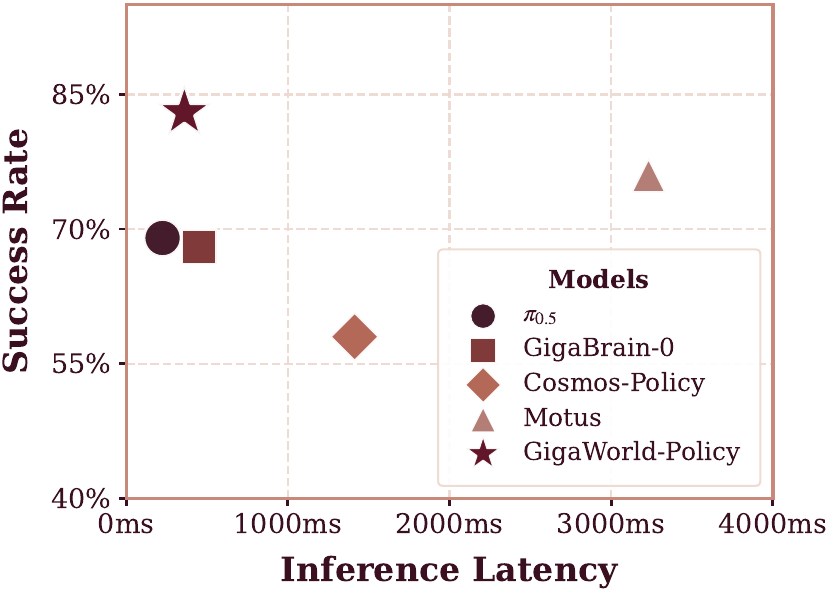}
    \caption{
        Comparison of GigaWorld-Policy with baselines on inference frequency and task success rate in real-world settings and on an A100 GPU.
    }
    \label{compare}
    \vspace{-4mm}
\end{wrapfigure}
Vision–Language–Action (VLA) models~\citep{pi_0,pi05,gr3,galaxea,gigabrain0,swiftvla,worldvla,dreamvla,vlar1} based on Vision–Language Models (VLMs)~\citep{paligemma,paligemma2,smolvlm} have achieved strong performance. However, a major challenge remains: supervision sparsity. While observations and task conditioning are high-dimensional and semantically rich, action supervision is sparse and low-diversity. This can lead the model to depend on shallow contextual signals, compressing varied situations into a few recurring behaviors rather than modeling physically grounded actions.

\begin{figure*}[t]
    \centering
    \captionsetup{type=figure, justification=justified, singlelinecheck=false}
    \includegraphics[width=.99\linewidth]{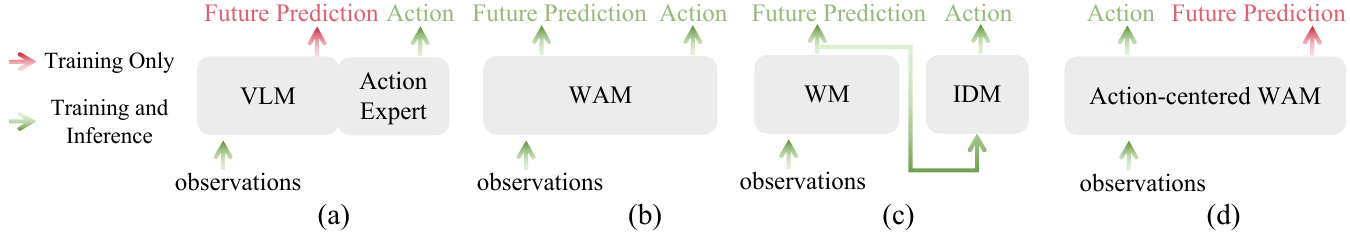} 
    \vspace{-3mm}
\caption{ (a) VLM-based VLA with auxiliary future supervision to densify training signals, but limited by the discriminative nature of VLMs~\citep{worldvla,dreamvla,swiftvla}. (b) Joint action--video prediction with  bidirectional attention~\citep{motus,cosmospolicy,shen2025videovla}, requiring future video generation at inference. (c) Two-stage design~\citep{mimicvideo,lingbotva} that first generates future video and then gets actions via an Inverse Dynamics Model (IDM). This approach still requires full video prediction to obtain actions, incurring additional inference overhead. (d) GigaWorld-Policy: an action-centered WAM that leverages future visual dynamics as supervision during training, while making future-video prediction optional at inference for low-latency action generation.
}
    \label{intro}
    \vspace{-4mm}
\end{figure*}

Therefore, some works~\citep{worldvla,dreamvla,swiftvla,chang2025scalable} attempt to inject future-state supervision into existing VLA frameworks by predicting future visual observations, as illustrated in Fig.~\ref{intro}~(a). However, VLM-based VLA~\citep{11357220,team2026gigabrain05,wang2025flowram,tian2025pdfactor,ding2025humanoid,cui2025openhelix,ding2024quar,wang2025vlaadapter,li2025vla} models are typically optimized for discriminative reasoning rather than high-fidelity generation, making it non-trivial for these additional losses to enforce continuity and physical consistency in the predicted actions.

In contrast, recent efforts incorporate the World Model (WM) from video generation~\citep{liuarflow,humandreamer,SEExiao,sdxl} into robot policy learning~\citep{motus,cosmospolicy,shen2025videovla,lingbotva,ye2026worldactionmodelszeroshot} to further increase supervision density and improve scalability.  Leveraging video generation is appealing because it provides temporally dense supervision in the observation space beyond sparse action labels and injects strong spatiotemporal priors learned from large-scale video data. These methods commonly optimize joint objectives of future visual dynamics and action prediction, explicitly coupling future observation forecasting with action selection, thereby leveraging the representational and generative capacity of video models to guide action learning (Fig.~\ref{intro}~(b–c)). However, these approaches often require iterative sampling to roll out future videos at inference time, leading to high latency. Moreover, errors in video prediction can propagate to action decoding, causing mistakes and degraded long-horizon control, particularly when small early inaccuracies compound over time.
To address these limitations, we introduce GigaWorld-Policy, an action-centered and efficient World--Action model. Instead of making action prediction overly reliant on explicit video generation, GigaWorld-Policy leverages future visual dynamics as a reasoning signal and a source of dense supervision. Specifically, GigaWorld-Policy is implemented as a causal sequence model that  represents action tokens and future-visual tokens under a causal mask. During training, the model learns to predict future action sequences from the current observation context, and in parallel learns an action-conditioned visual dynamics model that forecasts future visual observations given the same current observation together with the predicted actions, thereby coupling action learning with explicit 2D pixel-level state evolution. These two learning signals are optimized within the same model, allowing future visual dynamics to regularize action plausibility and provide substantially denser supervision, which improves learning efficiency. Crucially, at inference time, explicit future-video prediction is optional: the model can be executed in an action-only mode that directly produces control commands without rolling out long sequences of video tokens. This design substantially reduces compute and memory overhead, avoids compounding errors from extended visual rollouts, 
and enables low-latency closed-loop control, as illustrated in Fig.~\ref{intro}~(d). To obtain stronger pre-trained weights, we use a curriculum training pipeline that injects physics priors from diverse video sources before any task-specific supervision. GigaWorld-Policy is initialized from a large-scale web-video foundation model~\citep{wan}, and then further pre-trained on embodied, robot-centric data that combines real robot recordings with large-scale egocentric human videos, improving robustness to embodiment-specific viewpoints and interaction dynamics. Finally, we post-train the model on target-robot trajectories that align images, language, and actions, specializing it for instruction-conditioned action prediction under the target robot’s control interface and state distribution.

We validate GigaWorld-Policy through experiments in both simulation and real-world environments. GigaWorld-Policy outperforms strong baselines, delivering efficiency gains without sacrificing control performance. As shown in Fig.~\ref{compare}, it achieves a trade-off between success and efficiency, delivering $9\times$ faster inference and $7\%$ higher task success rates than the state-of-the-art WAM Motus~\citep{motus}. Moreover, under comparable speed and VLA settings, GigaWorld-Policy improves performance by 20\%.

The main contributions of this paper are summarized as follows:
\begin{itemize}[leftmargin=*,itemsep=2pt,topsep=2pt]
    \item We propose GigaWorld-Policy, an action-centered and efficient World--Action model. During training, future visual dynamics provide dense supervision and a reasoning signal for action learning, without over-reliance on explicit video synthesis. At inference time, future-visual prediction is optional and we can decode actions only, enabling low-latency control.
    \item We propose a pre-training paradigm that converts a generic video generation model into a strong initialization for robot policy learning, fully leveraging complementary data sources across stages.
    \item Experiments on real robotic platforms show that GigaWorld-Policy achieves a $9\times$ inference speedup (down to $0.36\,\mathrm{s}$ per inference) while improving task success rates by $7\%$ over baseline methods; compared with $\pi_{0.5}$, GigaWorld-Policy matches the inference speed and improves performance by $95\%$ on RoboTwin~2.0.
\end{itemize}

\section{Related Work}

\subsection{World Models for Robotic Video Generation}
Recent advances in world models~\citep{recondreamer,drivedreamer4d,recondreamer++,humandreamer,humandreamerx,Recondreamer-rl,mimicdreamer} have  improved robotic video generation and prediction~\citep{robotransfer,emma,robodreamer,ni2025wonderfree,embodiedreamer}. The central goal is to learn a generative model that captures the temporal evolution of the environment, enabling the prediction of future visual sequences.

Pandora~\citep{xiang2024pandora} proposes a hybrid autoregressive–diffusion world model that generates videos while enabling real-time control via free-text actions. FreeAction~\citep{kim2025freeaction} explicitly exploits continuous action parameters in diffusion-based robot video generation by using action-scaled classifier-free guidance to better control motion intensity. GigaWorld-0-video~\citep{gigaworld0} is a high-fidelity world-model data engine that synthesizes temporally coherent, high-quality 2D video sequences with fine-grained control over appearance and scene layout.  Some methods~\citep{chen2025visrl,chen2025think} also explore explicit video world models, which aim to construct structured and manipulable 3D scene representations~\citep{wang2025drivegen3d,wonderturbo,liu2024point,liu2025mamba4d}. Aether~\citep{team2025aether} unifies geometry-aware world modeling by jointly optimizing 4D dynamic reconstruction, action-conditioned video prediction, and goal-conditioned visual planning. 

However, most existing efforts improve the fidelity, consistency, and controllability of video world models, while largely overlooking how to adapt generic video generators into action-centered models that directly support policy learning under tight latency constraints. In contrast, we treat the video generator as policy initialization and propose an action-centered training recipe that aligns the backbone with robotic observations and action conditioned dynamics.

\subsection{World--Action Models for Robotic Control}
World--Action Models (WAM)~\citep{cosmospolicy,motus,shen2025videovla,wang2025sampo,lingbotva,ye2026worldactionmodelszeroshot}, grounded in the video generation paradigm, aim to predict robot actions and future visual dynamics within a unified framework. By modeling action-conditioned future observations, WAMs provide dense temporal supervision and a learned predictive prior that regularizes policy learning. 

As shown in Fig.~\ref{intro}~(b), VideoVLA~\citep{shen2025videovla} directly utilizes video generation models as pre-trained weights to explore the transformation of large-scale video generation models into robotic manipulation models, employing multimodal diffusion Transformers to jointly model video, language, and action modalities, achieving dual prediction of actions and future visual outcomes. Motus~\citep{motus} proposes a unified world model that leverages existing general pre-trained models and rich, shareable motion information, introducing a Mixture-of-Transformer (MoT) architecture to integrate three expert modules and adopting a UniDiffuser-style scheduler to enable flexible switching between different modeling modes. 

In contrast, as shown in Fig.~\ref{intro}~(c), Mimic-video~\citep{mimicvideo} adopts a two-stage pipeline: it first leverages an Internet-scale pre-trained video backbone to predict future visual observations, and then uses a flow-matching inverse-dynamics action decoder to map the resulting video latents into low-level robot actions. LingBot-VA~\citep{lingbotva} further introduces an autoregressive video-action world modeling framework that predicts the next world state while jointly generating the corresponding action sequence. It also adopts a closed-loop rolling prediction mechanism, updating future predictions with the latest real observations after each action segment to reduce error accumulation and trajectory drift.

However, these methods all typically require iterative diffusion sampling to generate future videos during inference, which introduces significant latency and limits real-time deployment. Meanwhile, an over-reliance on explicit video prediction can be fragile: pixel-level forecasting is sensitive to stochasticity observability, and small visual prediction errors may compound over long horizons, weakening the usefulness of the learned dynamics for robust action generation.

\begin{figure*}[t]
    \centering
    \includegraphics[width=0.99\linewidth]{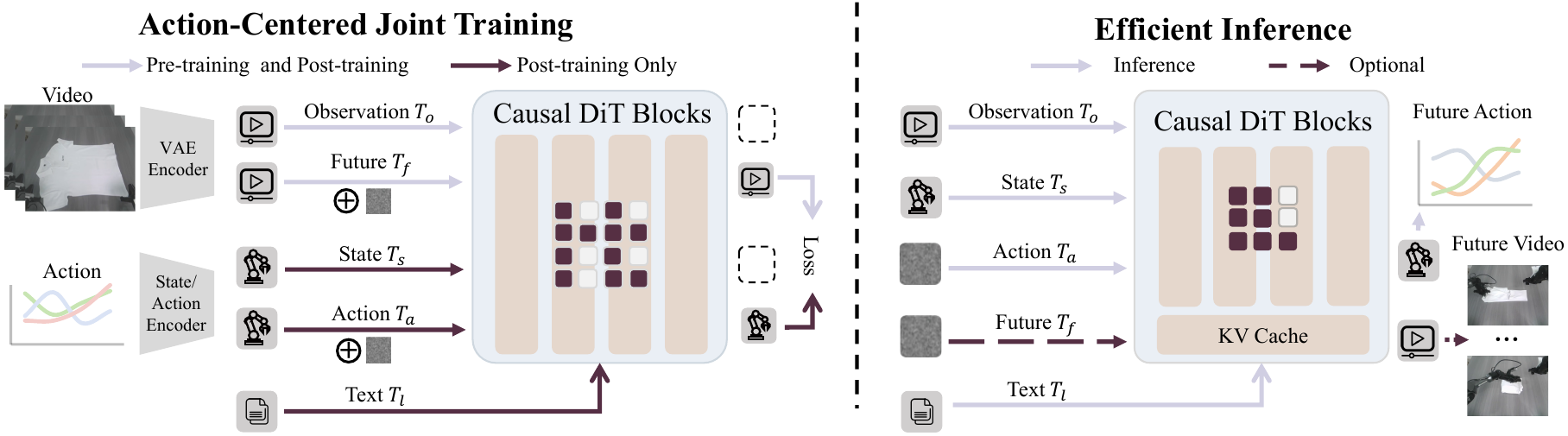} 
\caption{Overview of GigaWorld-Policy, built on a pre-trained video generation backbone. During pre-training, the model learns action-relevant representations from large-scale videos. During post-training for policy learning, it jointly performs action-chunk prediction from the current observation and future video prediction as auxiliary supervision. At inference time, the future-video prediction branch is optional, enabling faster inference.}
    \label{pipeline}
    \vspace{-4mm}
\end{figure*}

\section{Method}
\subsection{Problem Statement and Approach Overview}
We formulate robotic manipulation as a sequential decision-making task. At each time step $t$, the robot receives multi-view RGB observations $o_t=\{o_t^v\}_{v\in S}$ from a fixed set of camera viewpoints  $S=\{left,front,right\}$, a natural-language instruction $l$, and the proprioceptive state $s_t$. Conditioned on these inputs, the policy predicts an action chunk of length $p$, $a_{t:t+p-1}=(a_t,a_{t+1},\ldots,a_{t+p-1})$.

\noindent
\textbf{Vision-Language-Action Policies.} Most existing VLA policies are trained via imitation learning to model and sample an action chunk conditioned on the observation, the robot state, and a language instruction:
\begin{equation}
a_{t:t+p-1} \sim q_\Theta(\,\cdot \mid o_{t},\, s_t,\, l\,).
\end{equation}
The distribution $q_\Theta(\,\cdot \mid o_{ t}, s_t, l)$ parameterizes the policy over the next $p$ actions. In this paradigm, learning is driven solely by action supervision from demonstrations, without any explicit supervision in the observation space.

\noindent
\textbf{Our Approach.}
Unlike approaches that only model an action distribution, we adopt a world-modeling perspective that learns how visual observations evolve under an executed action chunk. We implement our method as a single unified model $g_\Theta$ that parameterizes two complementary conditional distributions. For action modeling, anchored by demonstrations, the model learns to sample an action chunk conditioned on the observation, robot state, and language:
\begin{equation}
\big(a_{t:t+p-1},\, c_t\big)\sim g_\Theta(\,\cdot \mid o_t,\, s_t,\, l\,),
\label{1}
\end{equation}
Here, $c_t$ is an action latent conditioning signal that is used to guide visual forecasting. For visual feedforward dynamics modeling, given the same context and the predicted action conditioning signal, the model learns to sample a future observation sequence that captures the evolution of visual observations:
\begin{equation}
(o_{t+\Delta},\, o_{t+2\Delta},\, \ldots,\, o_{t+K\Delta})
\sim g_\Theta\!\big(\,\cdot \mid o_t,\, s_t,\, l,\, c_t\big),
\label{2}
\end{equation}
where $\Delta$ denotes the temporal stride between predicted observations, and $K=\lfloor p/\Delta\rfloor$ so that the model predicts $\{o_{t+k\Delta}\}_{k=1}^{K}$ within the $p$-step horizon.




\subsection{The Architecture of GigaWorld-Policy}
As shown in Fig.~\ref{pipeline}, GigaWorld-Policy adapts a 5B-parameter diffusion Transformer~\citep{wan}, pre-trained via an action-centered objective to serve as a World--Action model for robotic manipulation. By concatenating multi-view inputs, the framework enables joint cross-view reasoning with consistency, and uses a causal masking scheme to unify action generation and visual dynamics.

\noindent
\textbf{Input Tokens.}
For visual token inputs, to enable multi-view generation without modifying the backbone while encouraging cross-view consistency, we merge the three camera views into a single composite image of the same resolution as a standard input:
\begin{equation}
o_t^{comp}={Compose}\!\left(o_t^{left},\,o_t^{front},\,o_t^{right}\right).
\label{concat}
\end{equation}
This composite representation preserves the spatial structure of each view in a shared coordinate frame, facilitating cross-view consistency. Meanwhile, since dense frame-by-frame prediction is frequently unnecessary due to strong temporal continuity and redundancy in adjacent observations, we only forecast a sparse set of future frames $\{\,o^{comp}_{t+k\Delta}\,\}_{k=1}^{K}$ using a fixed stride. Concretely, we predict one future observation every $\Delta$ steps along the action horizon, which preserves the key evolution of the scene while reducing supervision redundancy.

\noindent
\textbf{Shared Transformer Blocks.}
We then encode both the current observation $o_t^{comp}$ and the predicted future observations $\{\,o^{comp}_{t+k\Delta}\,\}_{k=1}^{K}$ using the same pre-trained Variational Autoencoder (VAE), and tokenize the resulting latents into spatiotemporal visual tokens $T_{o}$ and $T_{f}$. In parallel, we embed proprioceptive states and actions into the pre-trained model’s hidden dimension via linear projections, yielding state tokens $T_s$ and action tokens $T_a$, respectively. The language instruction $l$ is encoded by a pre-trained language encoder to obtain the instruction token sequence $T_{l}$.

Unlike MoE-based designs, we process all token types with a single shared stack of Transformer blocks.
In particular, all tokens share the same query, key, and value projection matrices at every layer, which tightly couples action tokens with visual evidence while preserving the computational profile of the pre-trained backbone. Meanwhile, we use different positional encodings for different token types to respect their underlying structures: visual tokens adopt a 2D positional encoding over the image grid, whereas proprioceptive state and action tokens use a 1D temporal positional encoding.

\begin{wrapfigure}{r}{0.35\textwidth} 
    \centering
    \includegraphics[width=0.99\linewidth]{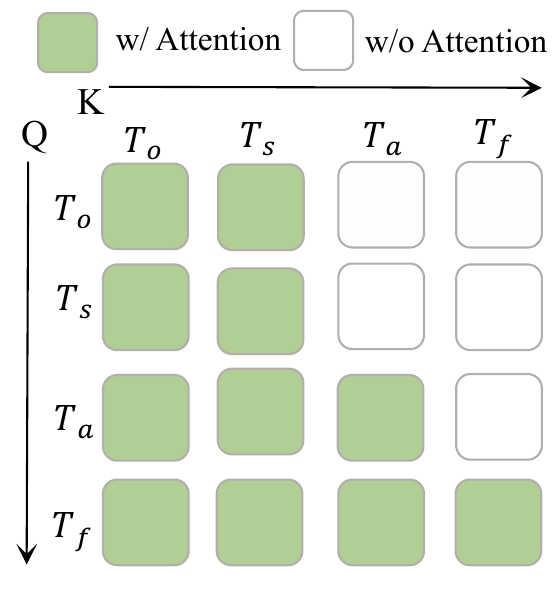}
\caption{Attention mask for GigaWorld-Policy: action tokens $T_{a}$ attend to states $T_{s}$ and current observations $T_{o}$ only, while future video tokens $T_{f}$ also attend to actions.}
    \vspace{-5mm} 
    \label{fig:mask}
\end{wrapfigure}
\noindent
\textbf{Causal Self-Attention for Video and Action Modeling.}
To unify action generation and feedforward visual-dynamics modeling within a single diffusion Transformer, we pack all modalities into one token sequence and use a causal attention mask to control information flow. Concretely, at each diffusion step $t$, we concatenate modality-specific tokens into a unified sequence:
\begin{equation}
T_t
=
\big[\; T_{o} \ ; \,T_s  \,;\; T_a \,;\; T_{f}\,\big].
\label{tokens}
\end{equation}
As shown in Fig.~\ref{fig:mask}, we then impose a blockwise causal attention mask  to enforce the following dependencies:
(i) $T_s$ and $T_{o}$ may attend to each other, but cannot attend to $T_a$ or $T_{f}$;
(ii) action tokens in $T_a$ may attend to the  tokens $\{T_s,\,T_{o}\}$, but cannot attend to $T_{f}$;
(iii) future-video tokens in $T_{f}$ may attend to $\{T_s,\,T_{o},\,T_a\}$, enabling feedforward dynamics prediction conditioned on the action chunk. This masking scheme prevents information leakage from predicted future frames into action generation, while allowing future-frame prediction to leverage both observations and actions, consistent with Eq.~\ref{1} and Eq.~\ref{2}.

Notably, the language instruction is not included in the unified self-attention sequence; instead, $l$ is provided as an external conditioning signal via cross-attention, and therefore does not participate in the causal ordering above.

\subsection{GigaWorld-Policy: Training}

\noindent\textbf{Training Process and Data.}
We pre-train GigaWorld-Policy to progressively inject physics priors from diverse video sources, enabling the model to acquire generalizable visual-dynamics knowledge before any task-specific supervision.

We first initialize GigaWorld-Policy from a large-scale pretrained video model~\citep{wan}, trained on diverse web videos. Building on this foundation, we perform embodied data pre-training, where the model is further pre-trained on robot-centered video data spanning real-world robot videos and large-scale egocentric human videos. On the robot side, we aggregate real-world robot videos from multiple sources (e.g., Agibot~\citep{agibot}, RDT~\citep{liu2024rdt}, RoboMind~\citep{robomind}, ATARA~\citep{feng2025vidar}), which capture robot-specific imaging characteristics, embodiment and workspace constraints, and the distinctive visual patterns induced by arms and end-effectors during interaction. In parallel, we include large-scale egocentric human demonstration videos (e.g., EgoDex~\citep{egodex}, Ego4D~\citep{grauman2022ego4d}) to broaden coverage of everyday interaction primitives and long-horizon activity structure, improving robustness to diverse scenes, tools, and task contexts. Overall, as shown in Tab.~\ref{tab:datasets_hours}, we collect approximately $10{,}000$ hours of data, and apply unified cleaning, formatting, and sampling across sources to ensure quality and a controllable data distribution. This embodied stage adapts the representation to embodiment-specific viewpoints and manipulation-relevant interaction patterns, improving robustness to viewpoint-induced appearance variations.

\begin{table}[t]
\centering
\small
\setlength{\tabcolsep}{13pt}
\caption{Datasets and their estimated collection hours used in embodied data pre-training, covering egocentric human demonstrations and real-world robot manipulation videos.}
\label{tab:datasets_hours}
\resizebox{\linewidth}{!}{%
\begin{tabular}{@{}l c l c l c@{}}
\toprule
Dataset & Hours & Dataset & Hours & Dataset & Hours \\
\midrule
EgoDex~\citep{egodex}               & 800       & Agibot~\citep{agibot}          & 2{,}500 & EGO4D~\citep{grauman2022ego4d}                    & 3{,}500 \\
RoboMind~\citep{robomind}           & 300     & RDT~\citep{liu2024rdt}         & 25        & Open X-Embodiment~\citep{o2024open}               & 3{,}500     \\
DROID~\citep{droid}                 & 350       & ATARA~\citep{feng2025vidar}    & 10        & Something-Something V2~\citep{goyal2017something} & 200   \\
\bottomrule
\end{tabular}%
}
\vspace{-2mm}
\end{table}

After pre-training, the model is post-trained on target-robot task trajectory data that pairs images, language, and actions. This stage specializes the model to the target robot by learning instruction-conditioned action prediction under the robot's control interface and state distribution.

\noindent\textbf{Training Objective.}
We use flow matching to optimize both action prediction and visual feedforward dynamics modeling. For either modality $x$ (action tokens or future video tokens), we sample a flow time $s\sim\mathcal{U}(0,1)$ and noise $\epsilon\sim\mathcal{N}(0,I)$, and construct the interpolated noised variable
\begin{equation}
x^{(s)}=(1-s)\epsilon+s x,
\end{equation}
with target velocity:
\begin{equation}
\dot{x}^{(s)}=x-\epsilon.
\end{equation}
Let $z_{f}$ denote the VAE latents corresponding to the future observation tokens $T_{f}$. We train the model to predict the velocity field of future latents conditioned on history and the executed action chunk:
\begin{equation}
\mathcal{L}_{video}
=
\mathbb{E}_{s,\epsilon}
\Big[
\big\|
g_\Theta\!\big(z_{f}^{(s)},\, s \mid T_s,\,T_{o},\,T_a,\,T_{l}\big)
-
\dot{z}_{f}^{(s)}
\big\|^2
\Big].
\end{equation}
Similarly, letting $a$ denote the action-token representation for the action chunk $T_a$, we optimize an action flow-matching objective conditioned on history:
\begin{equation}
\mathcal{L}_{action}
=
\mathbb{E}_{s,\epsilon}
\Big[
\big\|
g_\Theta\!\big(a^{(s)},\, s \mid T_s,\,T_{o},\,T_{l}\big)
-
\dot{a}^{(s)}
\big\|^2
\Big].
\end{equation}
For pre-training, we optimize only the video flow-matching objective. For post-training, we combine the video and action objectives using scalar weights $\lambda_{video}$ and $\lambda_{action}$ to balance their contributions:
\begin{equation}
\mathcal{L}_{all}
=
\lambda_{video}\,\mathcal{L}_{video}
+
\lambda_{action}\,\mathcal{L}_{action}.
\end{equation}

\subsection{GigaWorld-Policy: Inference}
At inference time, our goal is to generate actions with low latency. Directly running the unified video--action diffusion Transformer would require sampling the future video-token stream at every control step, which is costly because video tokens are typically much longer than action tokens. Moreover, predicting future frames is not necessary for executing the policy. We therefore adopt action decoding and optionally decode video, preserving the same backbone and tokenization as in training while avoiding video-generation overhead.

We keep the same conditioning pipeline for language, observations, and proprioception. At each control step $t$, we first form the context from the instruction, the current proprioceptive state, and the multi-view observation (concatenated as in Eq.~\ref{concat}):
\begin{equation}
w_t=\big(T_{l},\ T_s,\ T_{o}\big).
\end{equation}
We then sample only the action chunk tokens $T_a$ from the learned action flow model conditioned on $w_t$, without instantiating any future video tokens. Concretely, we initialize $a^{(0)}\sim \mathcal{N}(0,I)$ and integrate the learned velocity field from $s=0$ to $s=1$:
\begin{equation}
\frac{d a^{(s)}}{d s}=g_\Theta\!\big(a^{(s)},\, s \mid w_t\big),\qquad s\in[0,1],
\end{equation}
obtaining $a^{(1)}$, which is then decoded to the continuous action chunk $\hat{a}_{t:t+p-1}$. Finally, we execute the action, observe the new state and images, and repeat the above procedure at the next control step. If future prediction is needed, we can enable the video branch at inference time. This can be done either by including the future video tokens in the input and denoising them jointly with the action tokens, or by reusing the KV cache saved during action denoising and then denoising the video tokens conditioned on that cached context.




\section{Experiment}
\textbf{Evaluation Metrics.}
We primarily report the Success Rate (SR) to evaluate task completion performance. In simulation, SR is a binary indicator: SR $=1$ if the task is completed and SR $=0$ otherwise. For real-world experiments, we adopt a graded evaluation protocol. Specifically, for the pick-and-place task, we assign a score of $0.5$ for a successful grasp and an additional $0.5$ for successfully placing the object at the designated target location.

\noindent
\textbf{Baselines.}
We compare GigaWorld-Policy against several state-of-the-art baselines spanning two dominant paradigms. For VLM-based VLA methods, we include $\mathbf{\pi}_{0.5}$~\citep{pi05}, GigaBrain-0~\citep{gigabrain0}, and X-VLA~\citep{zheng2025x}, which leverage large VLM backbones to align visual observations with language instructions and decode low-level control commands via an action head, serving as strong representatives of end-to-end perception-to-action pipelines. For WAM approaches, we evaluate Cosmos-Policy~\citep{cosmospolicy} and Motus~\citep{motus}, which explicitly learn environment dynamics in a latent space and utilize predictive rollouts to support decision making.

\noindent\textbf{Implementation Details.}
Wan~2.2~5B~\citep{wan} serves as the diffusion-Transformer backbone. We set the action chunk length to $p=48$, and adopt a future-observation stride of $\Delta=12$ to provide auxiliary visual-dynamics supervision by sparsely predicting future observations along the action horizon. During post-training, we balance objectives with loss weights $\lambda_{action}=5$ and $\lambda_{video}=1$, emphasizing action prediction while retaining the video-consistency regularizer.  When reporting inference-time latency, we use the action-only decoding path, where future-video decoding is disabled and the model outputs actions directly. For real-world evaluations, each method is tested with 20 trials per task; in each trial, the robot is allowed up to five attempts to finish the task under the same instruction. In simulation, we evaluate each task over 100 test episodes.

\begin{table*}[t]
\caption{Evaluation on RoboTwin 2.0 Simulation. The best results are marked in \best{bold}, and the second-best results are \secondbest{underlined}. GigaWorld-Policy achieves performance comparable to Motus while providing a $9\times$ inference speedup.}
  \centering
  \footnotesize
  \setlength{\tabcolsep}{5.5pt}
  \resizebox{\textwidth}{!}{%
  \begin{tabular}{>{\centering\arraybackslash}m{3.6cm} *{8}{>{\centering\arraybackslash}m{0.7cm}}}
    \toprule
    \multirow{2}{*}{\makecell[c]{Simulation Task}}
      & \multicolumn{2}{c}{$\pi_{0.5}$}
      & \multicolumn{2}{c}{X-VLA}
      & \multicolumn{2}{c}{Motus}
      & \multicolumn{2}{c}{Our} \\
    & Clean & Rand.
    & Clean & Rand.
    & Clean & Rand.
    & Clean & Rand. \\
    \midrule

Adjust Bottle
  & 0.79 & 0.83 & \best{1.00} & \secondbest{0.99} & \secondbest{0.89} & 0.93 & \best{1.00} & \best{1.00} \\
Place A2b Left
  & 0.62 & 0.60 & 0.48 & 0.49 & \secondbest{0.88} & \secondbest{0.79} & \best{0.94} & \best{0.88} \\
Place Bread Skillet
  & 0.38 & 0.46 & 0.77 & 0.67 & \secondbest{0.86} & \secondbest{0.83} & \best{0.94} & \best{0.90} \\
Place Cans Plasticbox
  & 0.40 & 0.47 & 0.97 & \secondbest{0.98} & \secondbest{0.98} & 0.94 & \best{1.00} & \best{1.00} \\
Place Fan
  & 0.25 & 0.36 & 0.80 & 0.75 & \secondbest{0.91} & \secondbest{0.87} & \best{0.92} & \best{0.94} \\
Place Mouse Pad
  & 0.21 & 0.26 & \secondbest{0.70} & \secondbest{0.70} & 0.66 & 0.68 & \best{0.88} & \best{0.90} \\
Place Object Basket
  & 0.43 & 0.36 & 0.44 & 0.39 & \secondbest{0.81} & \secondbest{0.87} & \best{0.90} & \best{0.92} \\
Place Object Stand
  & 0.74 & 0.65 & 0.86 & 0.88 & \secondbest{0.98} & \secondbest{0.97} & \best{1.00} & \best{0.98} \\
Rotate Qrcode
  & 0.47 & 0.56 & 0.34 & 0.33 & \secondbest{0.89} & \secondbest{0.73} & \best{0.90} & \best{0.84} \\
Shake Bottle
  & 0.91 & \best{1.00} & \secondbest{0.99} & \best{1.00} & \best{1.00} & \secondbest{0.97} & \best{1.00} & \best{1.00} \\
Stamp Seal
  & 0.36 & 0.23 & 0.76 & 0.82 & \secondbest{0.93} & \secondbest{0.92} & \best{0.96} & \best{0.98} \\
Dump Bin Bigbin
  & 0.30 & 0.42 & 0.79 & 0.77 & \best{0.95} & \secondbest{0.91} & \secondbest{0.92} & \best{1.00} \\
Handover Block
  & 0.18 & 0.19 & 0.73 & 0.37 & \best{0.86} & \secondbest{0.73} & \secondbest{0.80} & \best{0.80} \\

    \midrule
    \multicolumn{9}{c}{\dots\ (50+ tasks in total)} \\
    \midrule

Handover Mic
  & 0.28 & 0.18 & 0.00 & 0.00 & \best{0.78} & \secondbest{0.63} & \secondbest{0.72} & \best{0.72} \\
Move Stapler Pad
  & 0.16 & 0.18 & 0.78 & 0.73 & \secondbest{0.83} & \best{0.85} & \best{0.92} & \secondbest{0.82} \\
Open Laptop
  & 0.19 & 0.35 & 0.93 & \best{1.00} & \secondbest{0.95} & 0.91 & \best{0.96} & \secondbest{0.98} \\
Place A2b Right
  & 0.62 & 0.57 & 0.36 & 0.36 & \best{0.91} & \secondbest{0.87} & \secondbest{0.90} & \best{0.92} \\
Place Burger Fries
  & 0.66 & 0.70 & \secondbest{0.94} & 0.94 & \best{0.98} & \best{0.98} & \best{0.98} & \secondbest{0.96} \\
Place Container Plate
  & 0.71 & 0.78 & \secondbest{0.97} & 0.95 & \best{0.98} & \best{0.99} & \best{0.98} & \secondbest{0.96} \\
Press Stapler
  & 0.80 & 0.70 & 0.92 & \best{0.98} & \secondbest{0.93} & \best{0.98} & \best{0.96} & \secondbest{0.96} \\
Put Object Cabinet & 0.24 & 0.15 & 0.46 & 0.48 & \best{0.88} & \secondbest{0.71} & \secondbest{0.74} & \best{0.74} \\
Place Dual Shoes & 0.12 & 0.07 & 0.79 & \best{0.88} & \secondbest{0.93} & \secondbest{0.87} & \best{0.96} & 0.84 \\

Stack Blocks Two
  & 0.48 & 0.56 & \secondbest{0.92} & 0.87 & \best{1.00} & \best{0.98} & \best{1.00} & \secondbest{0.94} \\
Turn Switch
  & 0.05 & 0.06 & 0.40 & 0.61 & \best{0.84} & \secondbest{0.78} & \secondbest{0.82} & \best{0.84} \\

    \midrule
    \textbf{Average}
      & 0.43 & 0.44
      & 0.73 & 0.73
      & \best{0.89} & \best{0.87}
      & \secondbest{0.87} & \secondbest{0.85} \\
    \bottomrule
  \end{tabular}
  }
  \label{tab:robotwin}
  \vspace{-0.3cm}
\end{table*}

\subsection{Inference Speed Comparison}
\begin{wraptable}{r}{0.52\columnwidth}
\vspace{-11mm}
\caption{Inference latency on an NVIDIA A100, and Success Rate in simulation and real world. The best results are marked in \best{bold}, and the second-best results are \secondbest{underlined}.}
\label{tab:inference_time}
\centering
\setlength{\tabcolsep}{6pt}
\resizebox{\linewidth}{!}{%
\begin{tabular}{lccc}
\toprule
Method & Time (ms) & \makecell[c]{SR in\\ Simulation} & \makecell[c]{SR in \\ Real-World} \\
\midrule
\multicolumn{4}{c}{Vision--Language--Action Models} \\
\midrule
$\pi_{0.5}$~\citep{pi05} & \best{225} & 0.48 & 0.69 \\
GigaBrain-0~\citep{gigabrain0} & 452 & -- & 0.68 \\
\midrule
\multicolumn{4}{c}{World--Action Models} \\
\midrule
Motus~\citep{motus} & 3231 & \best{0.88} & \secondbest{0.76} \\
Cosmos-Policy~\citep{cosmospolicy} & 1413 & -- & 0.58 \\
Ours & \secondbest{360} & \secondbest{0.86} & \best{0.83} \\
\bottomrule
\end{tabular}%
}
\vspace{-4mm}
\end{wraptable}
To assess deployment efficiency, we benchmark per-step inference latency on an NVIDIA A100 GPU. As summarized in Tab.~\ref{tab:inference_time}, we report inference time together with success rates in simulation and in the real world. GigaWorld-Policy is substantially faster than prior World--Action models, achieving an approximately $9\times$ speedup over Motus~\citep{motus} while maintaining comparable simulation performance and improving real-world success. Compared with the strong VLA baselines, GigaWorld-Policy trades a modest increase in latency for markedly better simulation success and a large gain in real-world performance, while remaining close to real-time operation. Overall, these results suggest that GigaWorld-Policy strikes a favorable balance between world-model-based capability and practical inference efficiency. 

\subsection{Simulation Benchmark Experiments}
\noindent
\textbf{Simulation Setup.} We evaluate single-task performance on RoboTwin 2.0~\citep{robotwin} over 50 representative manipulation tasks under domain randomization. To assess cross-task generalization and the effect of pre-training, we train all baselines and GigaWorld-Policy on a mixed dataset of 2,500 demonstrations from clean scenes (50 per task) and 25,000 demonstrations from heavily randomized scenes (500 per task). Randomization includes backgrounds, table clutter, table height perturbations, and lighting changes, testing robustness to distribution shifts.

\noindent
\textbf{Results.}
As shown in Tab.~\ref{tab:robotwin}, GigaWorld-Policy improves the average success rate by over $44$ percentage points compared with the $\pi_{0.5}$~\citep{pi05} model in the RoboTwin~2.0~\citep{robotwin} multi-task setting. Despite achieving an $9\times$ inference speedup, GigaWorld-Policy attains performance comparable to Motus~\citep{motus}. These results suggest that our action-centered World--Action modeling paradigm strengthens policy learning via joint supervision, combining action prediction with a visual-dynamics consistency auxiliary constraint during training to better align learned representations with downstream control. At inference time, GigaWorld-Policy makes future-video prediction optional, enabling low-latency action generation and largely accelerating inference without sacrificing overall task performance.

\begin{table*}[t]
\caption{Comparison of task success rates in real-world experiments. The best results are marked in \best{bold}, and the second-best results are \secondbest{underlined}.}
\centering
\setlength{\abovecaptionskip}{0.5em}
\setlength{\tabcolsep}{8pt}
\resizebox{0.99\linewidth}{!}{
\begin{tabular}{lccccc}  
\toprule
Method & \makecell{Clean the \\ Desk} & \makecell{Scan a\\QR Code} & \makecell{Sweep up \\ Trash} & \makecell{Stack Bowls} & Average \\
\midrule
\multicolumn{6}{c}{Vision--Language--Action Models} \\
\midrule
$\pi_{0.5}$~\citep{pi05} & 0.75 & 0.55 & 0.65 & \secondbest{0.80} & 0.69 \\
GigaBrain-0~\citep{gigabrain0} & 0.70 & \secondbest{0.65} & 0.60 & 0.75 & 0.68 \\
\midrule
\multicolumn{6}{c}{World--Action Models} \\
\midrule
Motus~\citep{motus} & \secondbest{0.80} & \best{0.75} & \secondbest{0.70} & \secondbest{0.80} & \secondbest{0.76} \\
Cosmos-Policy~\citep{cosmospolicy} & 0.65 & 0.50 & 0.45 & 0.70 & 0.58 \\
Our & \best{0.90} & \best{0.75} & \best{0.75} & \best{0.90} & \best{0.83} \\
\bottomrule
\end{tabular}
}
\label{tab:real}
\vspace{-1.5mm}
\end{table*}

\begin{figure*}[t]
    \centering
    \includegraphics[width=0.99\linewidth]{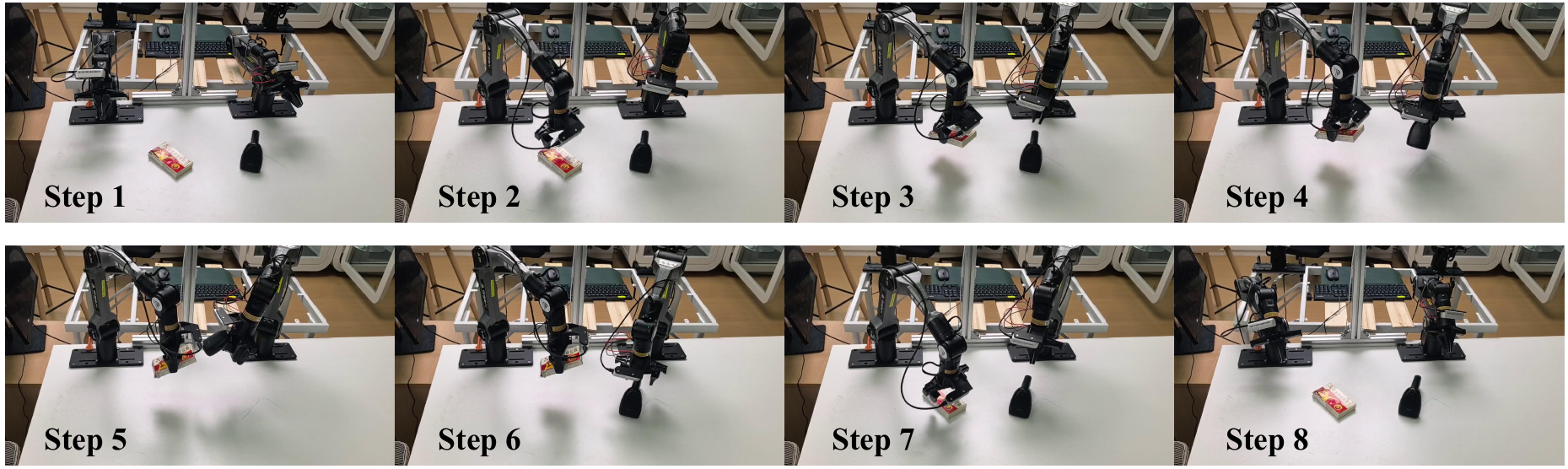} 
\caption{Real-world deployment of GigaWorld-Policy on PiPER arms for QR code scanning.}
    \label{task1}
\end{figure*}
\begin{figure*}[t]
    \centering
    \includegraphics[width=0.99\linewidth]{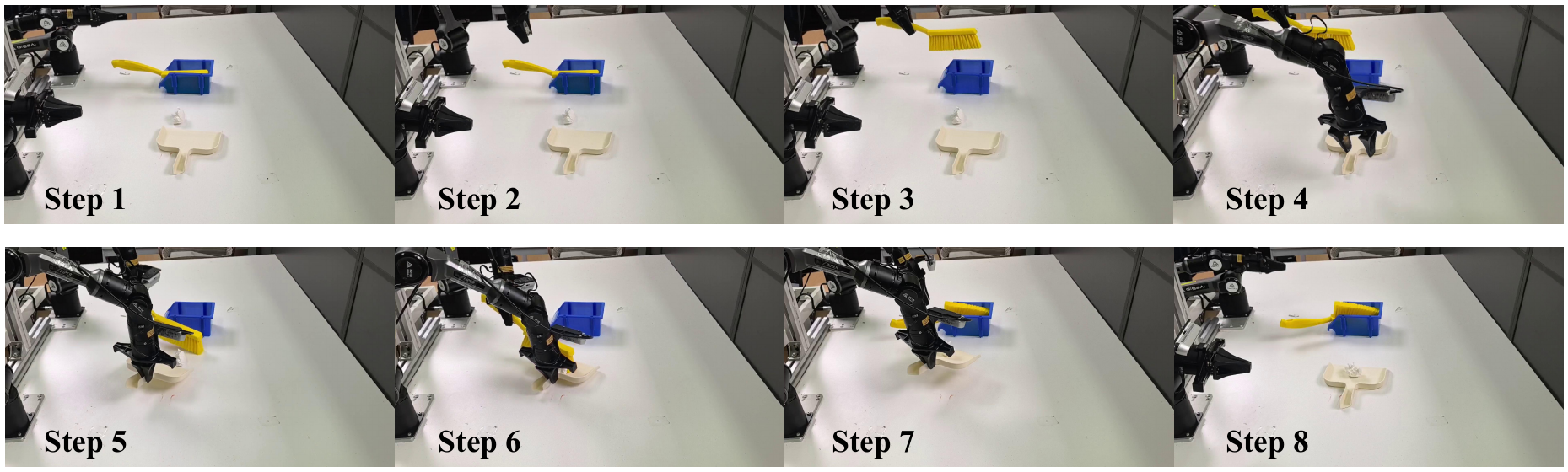} 
\caption{Real-world deployment of GigaWorld-Policy on PiPER arms for trash sweeping.}
    \label{task2}
\end{figure*}

\subsection{Real-World Experiment}
To evaluate real-world effectiveness, we conducted gripper-based manipulation experiments on an AgileX PiPER 6-DoF robotic arm. We further designed four real-world tasks—Clean the Desk, Scan a QR Code, Stack Bowls, and Sweep up Trash—with detailed descriptions provided in the supplementary material. As summarized in Tab.~\ref{tab:real}, GigaWorld-Policy achieves stronger overall performance than representative world-model-based robot policies. Notably, although GigaWorld-Policy runs nine times faster than Motus~\citep{motus}, it still attains a $7\%$ higher success rate. We attribute this advantage to the tight coupling between success and execution efficiency in real deployments: slower inference introduces control latency, reduces the effective action update rate, and weakens closed-loop error correction. GigaWorld-Policy also maintains a clear advantage over VLA baselines: compared with $\pi_{0.5}$~\citep{pi05}, it improves the average real-world success rate by about $14\%$ and achieves consistently higher success across all four tasks.

\subsection{Data Efficiency}
\begin{figure}[t]
    \centering
    \begin{minipage}{0.51\linewidth}
        \centering
        \includegraphics[width=\linewidth]{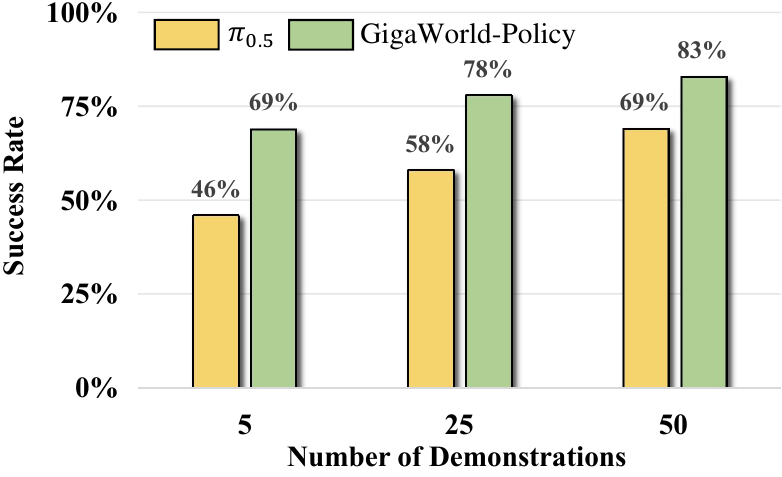}
        \captionof{figure}{Success rate as a function of training data fraction on real-world tasks. GigaWorld-Policy matches the $\pi_{0.5}$ with only $10\%$ of the data.}
        \label{fig:quantity3}
    \end{minipage}\hfill%
    \begin{minipage}{0.46\linewidth}
        \centering
        \includegraphics[width=1\linewidth]{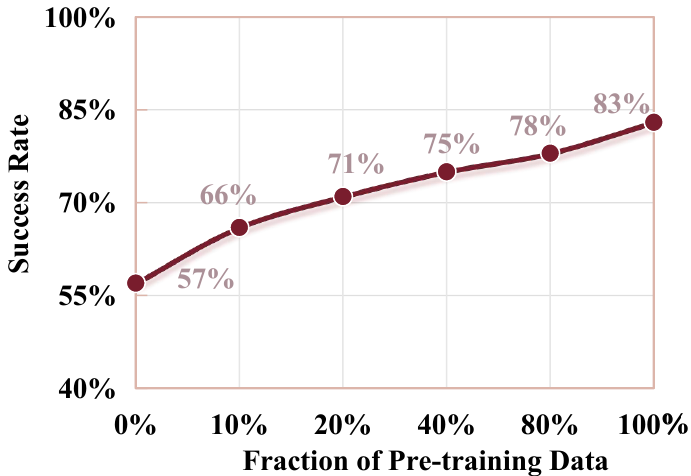}
       \caption{Increasing the fraction of embodied data pre-training consistently improves real-world success.}
        \label{fig:traindata}
    \end{minipage}
\end{figure}

Collecting real-world robot data is expensive, making sample-efficient policy learning critical. We study the data efficiency of GigaWorld-Policy and compare it against a VLA baseline. For each real-world task, we subsample the training demonstrations to multiple fractions, train all methods under the same setting, and evaluate success rates using the same protocol. As shown in Fig.~\ref{fig:quantity3}, conditioning on the video-prior representations learned by our World--Action model yields an improvement in sample efficiency. Specifically, GigaWorld-Policy reaches the maximum success rate achieved by the VLA using only $10\%$ of the training data. 

\begin{table}[t]
\caption{Effect of the sampling interval $\Delta$ on real-world task Success Rate. The model predicts $K = \lfloor 48/\Delta \rfloor$ future frames. $K = 0$ indicates no future video prediction.  The best results are marked in \best{bold}, and the second-best results are \secondbest{underlined}.}
\centering
\setlength{\abovecaptionskip}{0.5em}
\setlength{\tabcolsep}{18pt}
\resizebox{0.99\linewidth}{!}{
\begin{tabular}{ccccccc}
\toprule
$\Delta$ & 0 & 4 & 8 & 12 & 24 & 48 \\
\midrule 
Success Rate $\uparrow$ & 0.60 & 0.76 & \secondbest{0.78} & \best{0.83} & 0.80 & 0.76  \\
\bottomrule
\end{tabular}
}
\label{tab:ffdm_ablation}
\vspace{-1.5mm}
\end{table}

\begin{table*}[t]
\centering
\begin{minipage}[t]{0.47\textwidth}
\centering
\caption{Ablation study on the causal self-attention mask. We report SR and generation quality on the real-world test set. The best results are \best{bold}.}
\label{tab:causal_ablation}
\setlength{\tabcolsep}{12pt}
\resizebox{0.99\linewidth}{!}{
\begin{tabular}{lccc}
\toprule
Method & SR $\uparrow$ & PSNR $\uparrow$ & SSIM $\uparrow$ \\
\midrule
Self-Attn & 0.81 & 27.87 & 0.892 \\
Ours & \textbf{0.83} & \textbf{28.41} & \textbf{0.901} \\
\bottomrule
\end{tabular}
}
\end{minipage}
\hfill
\begin{minipage}[t]{0.50\textwidth}
\centering
\caption{Effect of different pre-training configurations on SR.}
\label{tab:ablation_stage}
\vspace{-2mm}
\setlength{\tabcolsep}{12pt}
\renewcommand{\arraystretch}{1.1}
\resizebox{0.99\linewidth}{!}{
\begin{tabular}{lccr}
\toprule
Pretrained Init & Embodied Data Pre-training & SR  $\uparrow$  \\
\midrule
           &            & 0.45 \\
\checkmark &            & 0.57 \\
  &   \checkmark         & 0.73 \\
\checkmark & \checkmark & \textbf{0.83} \\
\bottomrule
\end{tabular}
}
\end{minipage}
\end{table*}

\subsection{Ablation Study}
In this section, we conduct real-world ablation studies to isolate the contribution of each component to task success and robustness.

\noindent
\textbf{Importance of Pre-training.} We analyze our pre-training pipeline by ablating its components. Specifically, we compare four post-training settings: (i) training from scratch without any pre-training, (ii) initializing from the pretrained video model without embodied data pre-training, (iii) using only embodied data pre-training (without video-model initialization), and (iv) initializing from the same pretrained video model after applying embodied data pre-training. As shown in Tab.~\ref{tab:ablation_stage}, training from scratch achieves an SR of 0.45. Pretrained video initialization improves SR to 0.57, while using only embodied data pre-training yields 0.73. Combining both delivers the best result, indicating that they provide complementary benefits.  

We also study how the amount of data used for embodied-data pre-training affects performance. We keep the pretrained video initialization and all post-training settings fixed, and vary the fraction of data in embodied data pre-training. As shown in Fig.~\ref{fig:traindata}, increasing the embodied data fraction steadily improves real-world success, from $57\%$ without embodied pre-training to $83\%$ with the full dataset. We observe a clear and consistent improvement as the embodied pre-training data scales up, suggesting that increasing embodied coverage further strengthens the learned action-relevant representations and translates into higher real-world success.

\noindent
\textbf{Impact of the Number of Predicted Future Frames.}
We study whether feed-forward dynamics modeling improves action understanding and execution. As shown in Tab.~\ref{tab:ffdm_ablation}, we fix the action chunk length to $48$ and vary the number of predicted future frames by adjusting the sampling interval $\Delta$. Under this setting, the model predicts
$K=\lfloor 48/\Delta \rfloor$ future frames. When $K=0$, the model performs no future video prediction and degenerates to an action decoder conditioned only on the current observation. The results demonstrate a consistent improvement in success rate once feed-forward dynamics modeling is enabled, with performance increasing from $0.65$ to $0.83$, yielding an absolute gain of $0.18$. This suggests that anticipating future observations provides useful temporal priors for decision making, leading to more accurate and stable action execution. Meanwhile, we find that predicting denser future videos does not yield additional gains. This suggests that, a moderate amount of future modeling is sufficient to provide useful information for action prediction, while overly long prediction horizons exhibit diminishing returns and may even degrade performance.

\begin{figure*}[t]
    \centering
    \includegraphics[width=0.99\linewidth]{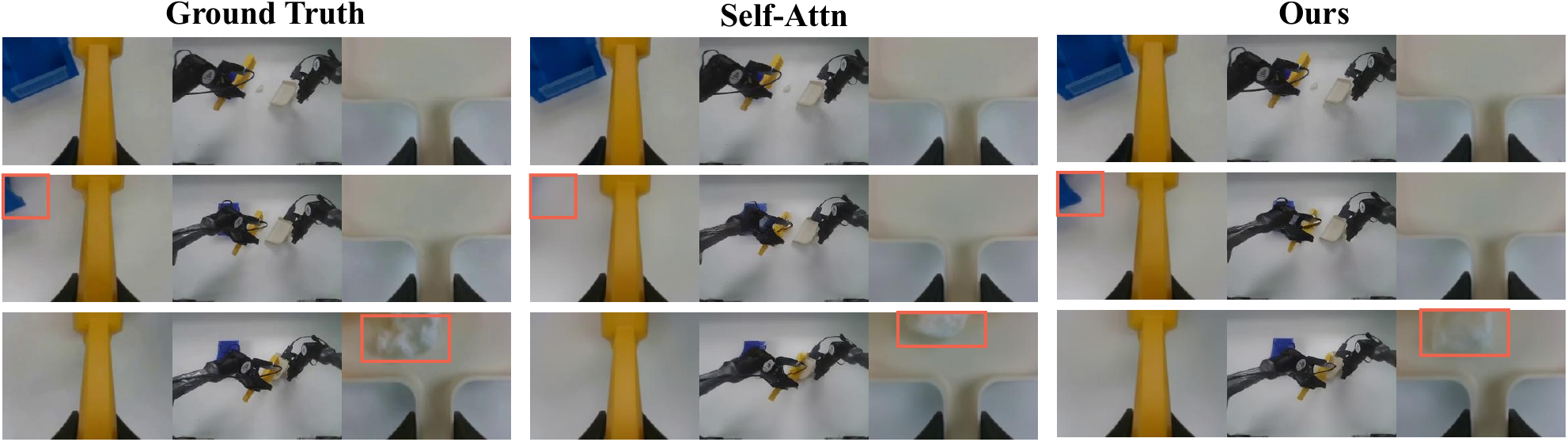} 
    \caption{Qualitative comparison between the self-attention baseline and our method. In the red boxes, our method more accurately predicts object state changes.}
    \label{vis}
    \vspace{-3mm}
\end{figure*}

\begin{table*}[t]
  \renewcommand{\arraystretch}{0.95}
\caption{Evaluation on RoboTwin 2.0 Simulation. The best results are marked in \best{bold}, and the second-best results are \secondbest{underlined}. GigaWorld-Policy achieves performance comparable to Motus while providing a $9\times$ inference speedup.}
  \centering
  \footnotesize
  \setlength{\tabcolsep}{5.5pt}
  \resizebox{\textwidth}{!}{
  \begin{tabular}{>{\centering\arraybackslash}m{4.2cm} *{8}{>{\centering\arraybackslash}m{0.7cm}}}
    \toprule
    \multirow{2}{*}{\makecell[c]{Simulation Task}}
      & \multicolumn{2}{c}{$\pi_{0.5}$}
      & \multicolumn{2}{c}{X-VLA}
      & \multicolumn{2}{c}{Motus}
      & \multicolumn{2}{c}{Our} \\
    & Clean & Rand.
    & Clean & Rand.
    & Clean & Rand.
    & Clean & Rand. \\
    \midrule

Adjust Bottle             & 0.79 & 0.83 & \best{1.00}  & \secondbest{0.99} & \secondbest{0.89} & 0.93          & \best{1.00}  & \best{1.00}  \\
Beat Block Hammer         & 0.63 & 0.50 & 0.92         & \secondbest{0.88} & \best{0.95}       & \best{0.88}   & \secondbest{0.86} & \secondbest{0.86} \\
Blocks Ranking Rgb        & 0.43 & 0.35 & 0.83         & 0.83              & \best{0.99}       & \best{0.97}   & \secondbest{0.92} & \secondbest{0.96} \\
Blocks Ranking Size       & 0.08 & 0.14 & \secondbest{0.67} & \best{0.74}  & \best{0.75}       & \secondbest{0.63} & 0.44         & 0.48          \\
Click Alarmclock          & 0.97 & 0.93 & 0.99         & 0.99              & \best{1.00}       & \best{1.00}   & \best{1.00}  & \best{1.00}   \\
Click Bell                & 0.75 & 0.76 & \best{1.00}  & \best{1.00}       & \best{1.00}       & \best{1.00}   & \best{1.00}  & \best{1.00}   \\
Dump Bin Bigbin           & 0.30 & 0.42 & 0.79         & 0.77              & \best{0.95}       & \secondbest{0.91} & \secondbest{0.92} & \best{1.00}  \\
Grab Roller               & 0.90 & 0.89 & \best{1.00}  & \best{1.00}       & \best{1.00}       & \best{1.00}   & \best{1.00}  & \best{1.00}   \\
Handover Block            & 0.18 & 0.19 & 0.73         & 0.37              & \best{0.86}       & \best{0.73}   & \secondbest{0.80} & \best{0.80}  \\
Handover Mic              & 0.28 & 0.18 & 0.00         & 0.00              & \best{0.78}       & \best{0.63}   & \secondbest{0.72} & \secondbest{0.72} \\
Hanging Mug               & 0.03 & 0.03 & \secondbest{0.23} & \secondbest{0.27} & \best{0.38}  & \best{0.38}   & 0.16         & 0.12          \\
Lift Pot                  & 0.00 & 0.00 & \best{0.99}  & \best{1.00}       & 0.96              & \secondbest{0.99} & \secondbest{0.98} & \secondbest{0.98} \\
Move Can Pot              & 0.29 & 0.27 & \best{0.89}  & \best{0.86}       & 0.34              & 0.74          & \secondbest{0.76} & \secondbest{0.78} \\
Move Pillbottle Pad       & 0.33 & 0.29 & 0.73         & 0.71              & \best{0.93}       & \best{0.96}   & \secondbest{0.90} & \secondbest{0.90} \\
Move Playingcard Away     & 0.59 & 0.67 & \secondbest{0.93} & \best{0.98}  & \best{1.00}       & \secondbest{0.96} & 0.78         & 0.72          \\
Move Stapler Pad          & 0.16 & 0.18 & 0.78         & 0.73              & \secondbest{0.83} & \best{0.85}   & \best{0.92}  & \secondbest{0.82} \\
Open Laptop               & 0.19 & 0.35 & 0.93         & \best{1.00}       & \secondbest{0.95} & 0.91          & \best{0.96}  & \secondbest{0.98} \\
Open Microwave            & 0.35 & 0.37 & \secondbest{0.79} & \secondbest{0.71} & \best{0.95}  & \best{0.91}   & 0.74         & 0.66          \\
Pick Diverse Bottles      & 0.05 & 0.03 & 0.58         & 0.36              & \best{0.90}       & \best{0.91}   & \secondbest{0.82} & \secondbest{0.70} \\
Pick Dual Bottles         & 0.10 & 0.06 & 0.47         & 0.36              & \best{0.96}       & \best{0.90}   & \secondbest{0.86} & \secondbest{0.86} \\
Place A2b Left            & 0.62 & 0.60 & 0.48         & 0.49              & \secondbest{0.88} & \secondbest{0.79} & \best{0.94}  & \best{0.88}  \\
Place A2b Right           & 0.62 & 0.57 & 0.36         & 0.36              & \secondbest{0.91} & \secondbest{0.87} & \best{0.90}  & \best{0.92}  \\
Place Bread Basket        & 0.48 & 0.56 & 0.81         & 0.71              & \secondbest{0.91} & \best{0.94}   & \secondbest{0.82} & \secondbest{0.82} \\
Place Bread Skillet       & 0.38 & 0.46 & 0.77         & 0.67              & \secondbest{0.86} & \secondbest{0.83} & \best{0.94}  & \best{0.90}  \\
Place Burger Fries        & 0.66 & 0.70 & 0.94         & 0.94              & \best{0.98}       & \best{0.98}   & \best{0.98}  & \secondbest{0.96} \\
Place Can Basket          & 0.19 & 0.25 & 0.49         & 0.52              & \best{0.81}       & \best{0.76}   & \secondbest{0.78} & \secondbest{0.74} \\
Place Cans Plasticbox     & 0.40 & 0.47 & \secondbest{0.97} & \secondbest{0.98} & \secondbest{0.98} & 0.94         & \best{1.00}  & \best{1.00}   \\
Place Container Plate     & 0.71 & 0.78 & 0.97         & 0.95              & \best{0.98}       & \secondbest{0.96} & \best{0.98}  & \best{0.96}  \\
Place Dual Shoes          & 0.12 & 0.07 & 0.79         & \secondbest{0.88} & \secondbest{0.93} & \best{0.87}   & \best{0.96}  & 0.84          \\
Place Empty Cup           & 0.75 & 0.86 & \best{1.00}  & \best{0.98}       & \secondbest{0.99} & \secondbest{0.98} & 0.90         & \secondbest{0.90} \\
Place Fan                 & 0.25 & 0.36 & 0.80         & 0.75              & \secondbest{0.91} & \secondbest{0.87} & \best{0.92}  & \best{0.94}  \\
Place Mouse Pad           & 0.21 & 0.26 & \secondbest{0.70} & \secondbest{0.70} & 0.66             & 0.68          & \best{0.88}  & \best{0.90}   \\
Place Object Basket       & 0.43 & 0.36 & 0.44         & 0.39              & \secondbest{0.81} & \secondbest{0.87} & \best{0.90}  & \best{0.92}  \\
Place Object Scale        & 0.40 & 0.49 & 0.52         & \secondbest{0.74} & \best{0.88}       & \secondbest{0.85} & \best{0.88}  & 0.80          \\
Place Object Stand        & 0.74 & 0.65 & 0.86         & 0.88              & \secondbest{0.98} & \secondbest{0.97} & \best{1.00}  & \best{0.98}  \\
Place Phone Stand         & 0.49 & 0.53 & \best{0.88}  & \best{0.87}       & \secondbest{0.87} & \secondbest{0.86} & 0.82         & 0.72          \\
Place Shoe                & 0.57 & 0.61 & 0.96         & 0.95              & \secondbest{0.99} & \secondbest{0.97} & \best{0.98}  & \best{0.96}  \\
Press Stapler             & 0.80 & 0.70 & 0.92         & \secondbest{0.98} & \secondbest{0.93} & \best{0.98}   & \best{0.96}  & \secondbest{0.96} \\
Put Bottles Dustbin       & 0.12 & 0.09 & \secondbest{0.74} & \best{0.77}  & \best{0.81}       & \secondbest{0.79} & 0.72         & 0.70          \\
Put Object Cabinet        & 0.24 & 0.15 & 0.46         & 0.48              & \best{0.88}       & \best{0.71}   & \secondbest{0.74} & \secondbest{0.74} \\
Rotate Qrcode             & 0.47 & 0.56 & 0.34         & 0.33              & \secondbest{0.89} & \best{0.73}   & \best{0.90}  & \secondbest{0.84} \\
Scan Object               & 0.42 & 0.38 & 0.14         & 0.36              & \best{0.67}       & \best{0.66}   & \secondbest{0.60} & \secondbest{0.64} \\
Shake Bottle Horizontally & 0.96 & \best{1.00} & \best{1.00} & \best{1.00}  & \best{1.00}       & \secondbest{0.98} & \best{1.00}  & \secondbest{0.98} \\
Shake Bottle              & 0.91 & \best{1.00} & 0.99 & \best{1.00}        & \best{1.00}       & \secondbest{0.97} & \best{1.00}  & \best{1.00}   \\
Stack Blocks Three        & 0.15 & 0.16 & 0.06         & 0.10              & \best{0.91}       & \best{0.95}   & \secondbest{0.70} & \secondbest{0.78} \\
Stack Blocks Two          & 0.48 & 0.56 & 0.92         & 0.87              & \secondbest{0.99} & \secondbest{0.98} & \best{1.00}  & \best{0.94}  \\
Stack Bowls Three         & 0.33 & 0.35 & \secondbest{0.76} & \best{0.86}  & \secondbest{0.79} & \secondbest{0.87} & 0.70         & 0.72          \\
Stack Bowls Two           & 0.78 & 0.66 & 0.96         & 0.93              & \best{0.98}       & \best{0.98}   & \secondbest{0.96} & \secondbest{0.92} \\
Stamp Seal                & 0.36 & 0.23 & 0.76         & 0.82              & \secondbest{0.93} & \secondbest{0.92} & \best{0.96}  & \best{0.98}  \\
Turn Switch               & 0.05 & 0.06 & 0.40         & \secondbest{0.61} & \best{0.84}       & \secondbest{0.78} & \secondbest{0.82} & \best{0.84}  \\
    \midrule
Average                   & 0.43 & 0.44 & 0.73       & 0.73             & \best{0.89}      & \best{0.87}  & \secondbest{0.86} & \secondbest{0.85} \\
    \bottomrule
  \end{tabular}
  }
  \label{robotwin1}
  \vspace{-0.3cm}
\end{table*}

\noindent
\textbf{Role of Causal Self-Attention for Video--Action Modeling.} 
We ablate the self-attention mask. Our causal design prevents leakage by restricting action tokens to attend only to the state and current observation. We compare this causal mask with a self-attention variant. In the self-attention variant, all tokens can attend to each other without constraints. As shown in Tab.~\ref{tab:causal_ablation}, both achieve similar SR under matched settings. However, causal self-attention is more deployable: it ensures the video prediction is optional at inference. We also compare future video generation quality on a held-out real-world test set from the target platform, using standard reconstruction metrics. As shown in Tab.\ref{tab:causal_ablation}, our method achieves higher PSNR and SSIM; furthermore, in Fig.\ref{vis}, the red boxed regions indicate that our method more accurately predicts fine-grained object state and appearance changes. We attribute this improvement to the proposed causal masking, which enforces a physically meaningful factorization between action prediction and forward visual dynamics. Under unconstrained self-attention, action tokens may attend to future-frame tokens during training, introducing information leakage that weakens action-conditioned dynamics learning. 

\section{Conclusion}
We introduced GigaWorld-Policy, an action-centered World--Action Model built on an action-conditioned video generation backbone pre-trained on a curated, multi-level large-scale robot dataset. By decomposing policy learning into future action sequence prediction together with action-conditioned video generation during training, GigaWorld-Policy learns 2D pixel--action dynamics with richer supervision, producing more accurate and plausible action plans. Experiments on real-world robotic platforms show that GigaWorld-Policy reduces inference overhead while improving task performance, achieving a $9\times$ inference speedup (down to $0.36\,\mathrm{s}$ per inference) and up to a $7\%$ increase in task success rates over prior baselines.

\appendix
\newpage
\makesuppltitle{}

\section{Implementation Details.}

\noindent
\textbf{Baselines.}
We compare SwiftWA with representative baselines from two dominant paradigms: VLM-based VLA methods, including $\pi_{0.5}$~\cite{pi05}, GigaBrain-0~\cite{gigabrain0}, and X-VLA~\cite{zheng2025x}, and World-Action Models, including Cosmos-Policy~\cite{cosmospolicy} and Motus~\cite{motus}.

$\pi_{0.5}$~\cite{pi05} is a vision-language-action (VLA) model built upon $\pi_0$ for open-world robotic generalization. It leverages co-training on heterogeneous tasks and diverse supervision sources, including data from multiple robots, high-level semantic prediction, web data, and low-level robot actions, enabling broader transfer to real-world manipulation scenarios. The model is initialized from a web-pretrained vision-language model and trained to combine high-level semantic reasoning with low-level control. Similar to $\pi_0$, the final policy uses Flow Matching to predict continuous action chunks for efficient real-time robot control.

X-VLA~\cite{zheng2025x} is a scalable cross-embodiment vision-language-action model that introduces embodiment-specific soft prompts to handle heterogeneity across diverse robotic data sources. Built on a shared Transformer backbone, it uses separate sets of learnable prompt embeddings for different robots or data domains, enabling efficient adaptation with minimal additional parameters while preserving generalization across embodiments. The model is designed for large-scale pretraining on heterogeneous robot datasets, and its final policy employs Flow Matching for continuous action prediction.

GigaBrain-0~\cite{gigabrain0} is a vision-language-action (VLA) foundation model that leverages world model-generated data to reduce reliance on large-scale real-world robot data. By incorporating diverse generated data, RGBD input modeling, and embodied Chain-of-Thought supervision, it improves cross-task generalization and robustness in dexterous, long-horizon, and mobile manipulation tasks.

Motus~\cite{motus} is a unified latent action world model that integrates understanding, video generation, and action modeling within a single Mixture-of-Transformer architecture. By learning latent actions from optical flow and supporting flexible mode switching through a UniDiffuser-style scheduler, it enables large-scale pretraining on heterogeneous data and achieves strong performance in both simulation and real-world robotic tasks.

Cosmos-Policy~\cite{cosmospolicy} is a robot policy that fine-tunes a pretrained video model for visuomotor control and planning in a single post-training stage without architectural changes. It directly generates actions, future state images, and value estimates as latent frames within the model's latent diffusion process, enabling both policy learning and test-time planning with strong performance in simulation and real-world manipulation tasks.

\noindent
\textbf{Pretraining Details.} We pretrain our model on public datasets for 6000 GPU hours using a global batch size of 256. We use the AdamW optimizer~\cite{kingma2014adam} with $\beta_{1}=0.85$ and $\beta_{2}=0.9$. The learning rate follows a cosine decay schedule, with an initial value of $1\times10^{-4}$ and a final value of $1\times10^{-6}$.

\begin{figure*}[t]
    \centering
    \includegraphics[width=0.99\linewidth]{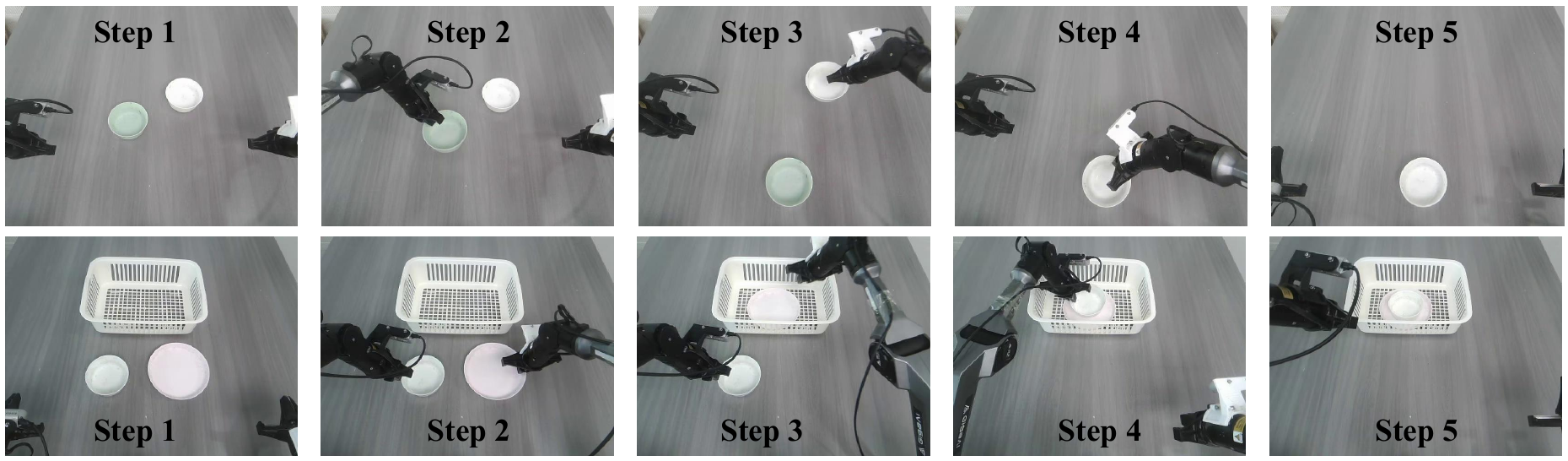} 
\caption{Real-world deployment of GigaWorld-Policy on PiPER arms for stacking bowls and cleaning a desk.}
    \label{task3}
    \vspace{-3mm}
\end{figure*}

\section{More details about the simulation experiments.}
Tab.~\ref{robotwin1} summarizes the evaluation results on the RoboTwin 2.0 simulation benchmark, where Motus and other baselines are tested on all 50 tasks under both clean and randomized scenes. This setup enables a thorough comparison of their effectiveness in standard conditions and their robustness when facing changes in scene configurations.

\section{More details about the real world experiments.}
As shown in Fig.~\ref{task1}, Fig.~\ref{task2} and Fig.~\ref{task3}, we illustrate the real-world scenarios and tasks used in our experiments, covering four manipulation tasks. For each task, we collect 50 demonstration trajectories, which are then used for post-training. The tasks are described as follows:

Clean the desk: The workspace contains a collection of bowls and plates with stochastically assigned color attributes, interspersed with randomly distributed obstacle objects. The robot is required to sequentially grasp and relocate all tableware items into a target basket, while satisfying a geometric ordering constraint that mandates plates be placed beneath bowls.

Stack bowls: Two bowls are initialized at arbitrary poses on the tabletop surface. The robot must identify and manipulate both objects to produce a properly nested configuration, with one bowl seated securely atop the other.

Scan a QR code: A QR code is affixed to a target object, which is placed at a randomized location within the workspace. The robot must first pick up a handheld scanner, then grasp the target object, align the scanner with the object's QR code to successfully read it, and finally return the object to its original location. This task evaluates the robot’s capabilities in multi-step task planning, tool use, and precise visual alignment for reliable decoding.

Sweep up Trash: Small objects are randomly scattered on a flat surface. The robot must first pick up a brush-like tool and a dustpan, then sweep the dispersed items into the dustpan using coordinated pushing motions. This task requires sustained contact force control and adaptive trajectory adjustment in response to the disordered initial configuration of the objects.

\clearpage

\setcitestyle{numbers}
\bibliographystyle{plainnat}
\bibliography{main}

\begin{thebibliography}{66}
\providecommand{\natexlab}[1]{#1}
\providecommand{\url}[1]{\texttt{#1}}
\expandafter\ifx\csname urlstyle\endcsname\relax
  \providecommand{\doi}[1]{doi: #1}\else
  \providecommand{\doi}{doi: \begingroup \urlstyle{rm}\Url}\fi

\bibitem[Beyer et~al.(2024)Beyer, Steiner, Pinto, Kolesnikov, Wang, Salz, Neumann, Alabdulmohsin, Tschannen, Bugliarello, et~al.]{paligemma}
Lucas Beyer, Andreas Steiner, Andr{\'e}~Susano Pinto, Alexander Kolesnikov, Xiao Wang, Daniel Salz, Maxim Neumann, Ibrahim Alabdulmohsin, Michael Tschannen, Emanuele Bugliarello, et~al.
\newblock Paligemma: A versatile 3b vlm for transfer.
\newblock \emph{arXiv preprint arXiv:2407.07726}, 2024.

\bibitem[Bi et~al.(2025)Bi, Tan, Xie, Wang, Huang, Liu, Zhao, Feng, Xiang, Rong, et~al.]{motus}
Hongzhe Bi, Hengkai Tan, Shenghao Xie, Zeyuan Wang, Shuhe Huang, Haitian Liu, Ruowen Zhao, Yao Feng, Chendong Xiang, Yinze Rong, et~al.
\newblock Motus: A unified latent action world model.
\newblock \emph{arXiv preprint arXiv:2512.13030}, 2025.

\bibitem[Black et~al.(2024)Black, Brown, Driess, Esmail, Equi, Finn, Fusai, Groom, Hausman, Ichter, et~al.]{pi_0}
Kevin Black, Noah Brown, Danny Driess, Adnan Esmail, Michael Equi, Chelsea Finn, Niccolo Fusai, Lachy Groom, Karol Hausman, Brian Ichter, et~al.
\newblock pi0: A vision-language-action flow model for general robot control.
\newblock \emph{arXiv preprint arXiv:2410.24164}, 2024.

\bibitem[Bu et~al.(2025)Bu, Cai, Chen, Cui, Ding, Feng, He, Huang, et~al.]{agibot}
Qingwen Bu, Jisong Cai, Li~Chen, Xiuqi Cui, Yan Ding, Siyuan Feng, Xindong He, Xu~Huang, et~al.
\newblock Agibot world colosseo: A large-scale manipulation platform for scalable and intelligent embodied systems.
\newblock In \emph{IROS}, 2025.

\bibitem[Cen et~al.(2025)Cen, Yu, Yuan, Jiang, Huang, Guo, Li, Song, Luo, Wang, et~al.]{worldvla}
Jun Cen, Chaohui Yu, Hangjie Yuan, Yuming Jiang, Siteng Huang, Jiayan Guo, Xin Li, Yibing Song, Hao Luo, Fan Wang, et~al.
\newblock Worldvla: Towards autoregressive action world model.
\newblock \emph{arXiv preprint arXiv:2506.21539}, 2025.

\bibitem[Chang et~al.(2025)Chang, Qin, Qiao, Wang, Zhu, Ma, and Wang]{chang2025scalable}
Yifan Chang, Jie Qin, Limeng Qiao, Xiaofeng Wang, Zheng Zhu, Lin Ma, and Xingang Wang.
\newblock Scalable training for vector-quantized networks with 100\% codebook utilization.
\newblock \emph{arXiv preprint arXiv:2509.10140}, 2025.

\bibitem[Cheang et~al.(2025)Cheang, Chen, Cui, Hu, Huang, Kong, Li, Li, Liu, Ma, et~al.]{gr3}
Chilam Cheang, Sijin Chen, Zhongren Cui, Yingdong Hu, Liqun Huang, Tao Kong, Hang Li, Yifeng Li, Yuxiao Liu, Xiao Ma, et~al.
\newblock Gr-3 technical report.
\newblock \emph{arXiv preprint arXiv:2507.15493}, 2025.

\bibitem[Chen et~al.(2025{\natexlab{a}})Chen, Chen, Chen, Cai, Liu, Liang, Li, Lin, Ge, Gu, et~al.]{robotwin}
Tianxing Chen, Zanxin Chen, Baijun Chen, Zijian Cai, Yibin Liu, Qiwei Liang, Zixuan Li, Xianliang Lin, Yiheng Ge, Zhenyu Gu, et~al.
\newblock Robotwin 2.0: A scalable data generator and benchmark with strong domain randomization for robust bimanual robotic manipulation.
\newblock \emph{arXiv preprint arXiv:2506.18088}, 2025{\natexlab{a}}.

\bibitem[Chen et~al.(2025{\natexlab{b}})Chen, Luo, and Li]{chen2025visrl}
Zhangquan Chen, Xufang Luo, and Dongsheng Li.
\newblock Visrl: Intention-driven visual perception via reinforced reasoning.
\newblock \emph{arXiv preprint arXiv:2503.07523}, 2025{\natexlab{b}}.

\bibitem[Chen et~al.(2025{\natexlab{c}})Chen, Zhang, Yu, Luo, Sun, Pan, Feng, Pei, Cai, and Huang]{chen2025think}
Zhangquan Chen, Manyuan Zhang, Xinlei Yu, Xufang Luo, Mingze Sun, Zihao Pan, Yan Feng, Peng Pei, Xunliang Cai, and Ruqi Huang.
\newblock Think with 3d: Geometric imagination grounded spatial reasoning from limited views.
\newblock \emph{arXiv preprint arXiv:2510.18632}, 2025{\natexlab{c}}.

\bibitem[Cui et~al.(2025)Cui, Ding, Song, Bai, Tong, Ge, Suo, Zhou, Liu, Jia, et~al.]{cui2025openhelix}
Can Cui, Pengxiang Ding, Wenxuan Song, Shuanghao Bai, Xinyang Tong, Zirui Ge, Runze Suo, Wanqi Zhou, Yang Liu, Bofang Jia, et~al.
\newblock Openhelix: A short survey, empirical analysis, and open-source dual-system vla model for robotic manipulation.
\newblock \emph{arXiv preprint arXiv:2505.03912}, 2025.

\bibitem[Ding et~al.(2024)Ding, Zhao, Zhang, Song, Zhang, Huang, Yang, and Wang]{ding2024quar}
Pengxiang Ding, Han Zhao, Wenjie Zhang, Wenxuan Song, Min Zhang, Siteng Huang, Ningxi Yang, and Donglin Wang.
\newblock Quar-vla: Vision-language-action model for quadruped robots.
\newblock In \emph{European Conference on Computer Vision}, pages 352--367. Springer, 2024.

\bibitem[Ding et~al.(2025)Ding, Ma, Tong, Zou, Luo, Fan, Wang, Lu, Mo, Liu, et~al.]{ding2025humanoid}
Pengxiang Ding, Jianfei Ma, Xinyang Tong, Binghong Zou, Xinxin Luo, Yiguo Fan, Ting Wang, Hongchao Lu, Panzhong Mo, Jinxin Liu, et~al.
\newblock Humanoid-vla: Towards universal humanoid control with visual integration.
\newblock \emph{arXiv preprint arXiv:2502.14795}, 2025.

\bibitem[Dong et~al.(2025)Dong, Wang, Zhu, Wang, Wang, Zhou, Wang, Ni, Ouyang, Qin, et~al.]{emma}
Zhehao Dong, Xiaofeng Wang, Zheng Zhu, Yirui Wang, Yang Wang, Yukun Zhou, Boyuan Wang, Chaojun Ni, Runqi Ouyang, Wenkang Qin, et~al.
\newblock Emma: Generalizing real-world robot manipulation via generative visual transfer.
\newblock \emph{arXiv preprint arXiv:2509.22407}, 2025.

\bibitem[Feng et~al.(2025)Feng, Tan, Mao, Xiang, Liu, Huang, Su, and Zhu]{feng2025vidar}
Yao Feng, Hengkai Tan, Xinyi Mao, Chendong Xiang, Guodong Liu, Shuhe Huang, Hang Su, and Jun Zhu.
\newblock Vidar: Embodied video diffusion model for generalist manipulation.
\newblock \emph{arXiv preprint arXiv:2507.12898}, 2025.

\bibitem[Goyal et~al.(2017)Goyal, Ebrahimi~Kahou, Michalski, Materzynska, Westphal, Kim, Haenel, Fruend, Yianilos, Mueller-Freitag, et~al.]{goyal2017something}
Raghav Goyal, Samira Ebrahimi~Kahou, Vincent Michalski, Joanna Materzynska, Susanne Westphal, Heuna Kim, Valentin Haenel, Ingo Fruend, Peter Yianilos, Moritz Mueller-Freitag, et~al.
\newblock The" something something" video database for learning and evaluating visual common sense.
\newblock In \emph{Proceedings of the IEEE international conference on computer vision}, pages 5842--5850, 2017.

\bibitem[Grauman et~al.(2022)Grauman, Westbury, Byrne, Chavis, Furnari, Girdhar, Hamburger, Jiang, Liu, Liu, et~al.]{grauman2022ego4d}
Kristen Grauman, Andrew Westbury, Eugene Byrne, Zachary Chavis, Antonino Furnari, Rohit Girdhar, Jackson Hamburger, Hao Jiang, Miao Liu, Xingyu Liu, et~al.
\newblock Ego4d: Around the world in 3,000 hours of egocentric video.
\newblock In \emph{Proceedings of the IEEE/CVF conference on computer vision and pattern recognition}, pages 18995--19012, 2022.

\bibitem[Hoque et~al.(2025)Hoque, Huang, Yoon, Sivapurapu, and Zhang]{egodex}
Ryan Hoque, Peide Huang, David~J Yoon, Mouli Sivapurapu, and Jian Zhang.
\newblock Egodex: Learning dexterous manipulation from large-scale egocentric video.
\newblock \emph{arXiv preprint arXiv:2505.11709}, 2025.

\bibitem[Intelligence et~al.(2025)Intelligence, Black, Brown, Darpinian, Dhabalia, Driess, Esmail, Equi, Finn, Fusai, et~al.]{pi05}
Physical Intelligence, Kevin Black, Noah Brown, James Darpinian, Karan Dhabalia, Danny Driess, Adnan Esmail, Michael Equi, Chelsea Finn, Niccolo Fusai, et~al.
\newblock $\pi_{0.5}$: a vision-language-action model with open-world generalization.
\newblock \emph{arXiv preprint arXiv:2504.16054}, 2025.

\bibitem[Jiang et~al.(2025)Jiang, Yuan, Liu, Lu, Cui, Liu, Cheng, Gao, Xu, and Zhao]{galaxea}
Tao Jiang, Tianyuan Yuan, Yicheng Liu, Chenhao Lu, Jianning Cui, Xiao Liu, Shuiqi Cheng, Jiyang Gao, Huazhe Xu, and Hang Zhao.
\newblock Galaxea open-world dataset and g0 dual-system vla model.
\newblock \emph{arXiv preprint arXiv:2509.00576}, 2025.

\bibitem[Khazatsky et~al.(2024)Khazatsky, Pertsch, Nair, Balakrishna, Dasari, Karamcheti, Nasiriany, Srirama, Chen, Ellis, et~al.]{droid}
Alexander Khazatsky, Karl Pertsch, Suraj Nair, Ashwin Balakrishna, Sudeep Dasari, Siddharth Karamcheti, Soroush Nasiriany, Mohan~Kumar Srirama, Lawrence~Yunliang Chen, Kirsty Ellis, et~al.
\newblock Droid: A large-scale in-the-wild robot manipulation dataset.
\newblock \emph{arXiv preprint arXiv:2403.12945}, 2024.

\bibitem[Kim et~al.(2026)Kim, Gao, Lin, Lin, Ge, Lam, Liang, Song, Liu, Finn, et~al.]{cosmospolicy}
Moo~Jin Kim, Yihuai Gao, Tsung-Yi Lin, Yen-Chen Lin, Yunhao Ge, Grace Lam, Percy Liang, Shuran Song, Ming-Yu Liu, Chelsea Finn, et~al.
\newblock Cosmos policy: Fine-tuning video models for visuomotor control and planning.
\newblock \emph{arXiv preprint arXiv:2601.16163}, 2026.

\bibitem[Kim et~al.(2025)Kim, Lee, and Cho]{kim2025freeaction}
Seungwook Kim, Seunghyeon Lee, and Minsu Cho.
\newblock Freeaction: Training-free techniques for enhanced fidelity of trajectory-to-video generation.
\newblock \emph{arXiv preprint arXiv:2509.24241}, 2025.

\bibitem[Li et~al.(2025{\natexlab{a}})Li, Zhang, Ouyang, Wang, Zhu, Yang, Zhang, Wang, Ni, Qin, et~al.]{mimicdreamer}
Haoyun Li, Ivan Zhang, Runqi Ouyang, Xiaofeng Wang, Zheng Zhu, Zhiqin Yang, Zhentao Zhang, Boyuan Wang, Chaojun Ni, Wenkang Qin, et~al.
\newblock Mimicdreamer: Aligning human and robot demonstrations for scalable vla training.
\newblock \emph{arXiv preprint arXiv:2509.22199}, 2025{\natexlab{a}}.

\bibitem[Li et~al.(2025{\natexlab{b}})Li, Ding, Suo, Wang, Ge, Zang, Yu, Sun, Zhang, Wang, et~al.]{li2025vla}
Hengtao Li, Pengxiang Ding, Runze Suo, Yihao Wang, Zirui Ge, Dongyuan Zang, Kexian Yu, Mingyang Sun, Hongyin Zhang, Donglin Wang, et~al.
\newblock Vla-rft: Vision-language-action reinforcement fine-tuning with verified rewards in world simulators.
\newblock \emph{arXiv preprint arXiv:2510.00406}, 2025{\natexlab{b}}.

\bibitem[Li et~al.(2026)Li, Zhang, Luo, Yang, Wang, Han, Yu, Gao, Xue, Zhu, Shen, and Xu]{lingbotva}
Lin Li, Qihang Zhang, Yiming Luo, Shuai Yang, Ruilin Wang, Fei Han, Mingrui Yu, Zelin Gao, Nan Xue, Xing Zhu, Yujun Shen, and Yinghao Xu.
\newblock Causal world modeling for robot control, 2026.

\bibitem[Liu et~al.(2024{\natexlab{a}})Liu, Yu, Wang, Zheng, Deng, Ye, and Wang]{liu2024point}
Jiuming Liu, Ruiji Yu, Yian Wang, Yu~Zheng, Tianchen Deng, Weicai Ye, and Hesheng Wang.
\newblock Point mamba: A novel point cloud backbone based on state space model with octree-based ordering strategy.
\newblock \emph{arXiv preprint arXiv:2403.06467}, 2024{\natexlab{a}}.

\bibitem[Liu et~al.(2025{\natexlab{a}})Liu, Han, Liu, Aviles-Rivero, Jiang, Liu, and Wang]{liu2025mamba4d}
Jiuming Liu, Jinru Han, Lihao Liu, Angelica~I Aviles-Rivero, Chaokang Jiang, Zhe Liu, and Hesheng Wang.
\newblock Mamba4d: Efficient 4d point cloud video understanding with disentangled spatial-temporal state space models.
\newblock In \emph{Proceedings of the Computer Vision and Pattern Recognition Conference}, pages 17626--17636, 2025{\natexlab{a}}.

\bibitem[Liu et~al.(2026)Liu, Liu, Zhu, Zhang, Li, Yang, Nex, Cheng, and Wang]{liuarflow}
Jiuming Liu, Mengmeng Liu, Siting Zhu, Yunpeng Zhang, Jiangtao Li, Michael~Ying Yang, Francesco Nex, Hao Cheng, and Hesheng Wang.
\newblock Arflow: Auto-regressive optical flow estimation for arbitrary-length videos via progressive next-frame forecasting.
\newblock In \emph{The Fourteenth International Conference on Learning Representations}, 2026.

\bibitem[Liu et~al.(2025{\natexlab{b}})Liu, Wang, Zhao, Li, Qin, Qiu, Zhu, Huang, and Su]{robotransfer}
Liu Liu, Xiaofeng Wang, Guosheng Zhao, Keyu Li, Wenkang Qin, Jiaxiong Qiu, Zheng Zhu, Guan Huang, and Zhizhong Su.
\newblock Robotransfer: Geometry-consistent video diffusion for robotic visual policy transfer.
\newblock \emph{arXiv preprint arXiv:2505.23171}, 2025{\natexlab{b}}.

\bibitem[Liu et~al.(2024{\natexlab{b}})Liu, Wu, Li, Tan, Chen, Wang, Xu, Su, and Zhu]{liu2024rdt}
Songming Liu, Lingxuan Wu, Bangguo Li, Hengkai Tan, Huayu Chen, Zhengyi Wang, Ke~Xu, Hang Su, and Jun Zhu.
\newblock Rdt-1b: a diffusion foundation model for bimanual manipulation.
\newblock \emph{arXiv preprint arXiv:2410.07864}, 2024{\natexlab{b}}.

\bibitem[Marafioti et~al.(2025)Marafioti, Zohar, Farr{\'e}, Noyan, Bakouch, Cuenca, Zakka, Allal, Lozhkov, Tazi, et~al.]{smolvlm}
Andr{\'e}s Marafioti, Orr Zohar, Miquel Farr{\'e}, Merve Noyan, Elie Bakouch, Pedro Cuenca, Cyril Zakka, Loubna~Ben Allal, Anton Lozhkov, Nouamane Tazi, et~al.
\newblock Smolvlm: Redefining small and efficient multimodal models.
\newblock \emph{arXiv preprint arXiv:2504.05299}, 2025.

\bibitem[Ni et~al.(2025{\natexlab{a}})Ni, Chen, Wang, Zhu, Zheng, Wang, Chen, Zhao, Li, Dong, et~al.]{swiftvla}
Chaojun Ni, Cheng Chen, Xiaofeng Wang, Zheng Zhu, Wenzhao Zheng, Boyuan Wang, Tianrun Chen, Guosheng Zhao, Haoyun Li, Zhehao Dong, et~al.
\newblock Swiftvla: Unlocking spatiotemporal dynamics for lightweight vla models at minimal overhead.
\newblock \emph{arXiv preprint arXiv:2512.00903}, 2025{\natexlab{a}}.

\bibitem[Ni et~al.(2025{\natexlab{b}})Ni, Li, Li, Liu, Wang, Zhu, Zhao, Wang, Li, Huang, et~al.]{ni2025wonderfree}
Chaojun Ni, Jie Li, Haoyun Li, Hengyu Liu, Xiaofeng Wang, Zheng Zhu, Guosheng Zhao, Boyuan Wang, Chenxin Li, Guan Huang, et~al.
\newblock Wonderfree: Enhancing novel view quality and cross-view consistency for 3d scene exploration.
\newblock \emph{arXiv preprint arXiv:2506.20590}, 2025{\natexlab{b}}.

\bibitem[Ni et~al.(2025{\natexlab{c}})Ni, Wang, Zhu, Wang, Li, Zhao, Li, Qin, Huang, and Mei]{wonderturbo}
Chaojun Ni, Xiaofeng Wang, Zheng Zhu, Weijie Wang, Haoyun Li, Guosheng Zhao, Jie Li, Wenkang Qin, Guan Huang, and Wenjun Mei.
\newblock Wonderturbo: Generating interactive 3d world in 0.72 seconds.
\newblock \emph{arXiv preprint arXiv:2504.02261}, 2025{\natexlab{c}}.

\bibitem[Ni et~al.(2025{\natexlab{d}})Ni, Zhao, Wang, Zhu, Qin, Chen, Jia, Huang, and Mei]{Recondreamer-rl}
Chaojun Ni, Guosheng Zhao, Xiaofeng Wang, Zheng Zhu, Wenkang Qin, Xinze Chen, Guanghong Jia, Guan Huang, and Wenjun Mei.
\newblock Recondreamer-rl: Enhancing reinforcement learning via diffusion-based scene reconstruction.
\newblock \emph{arXiv preprint arXiv:2508.08170}, 2025{\natexlab{d}}.

\bibitem[Ni et~al.(2025{\natexlab{e}})Ni, Zhao, Wang, Zhu, Qin, Huang, Liu, Chen, Wang, Zhang, et~al.]{recondreamer}
Chaojun Ni, Guosheng Zhao, Xiaofeng Wang, Zheng Zhu, Wenkang Qin, Guan Huang, Chen Liu, Yuyin Chen, Yida Wang, Xueyang Zhang, et~al.
\newblock Recondreamer: Crafting world models for driving scene reconstruction via online restoration.
\newblock In \emph{Proceedings of the Computer Vision and Pattern Recognition Conference}, pages 1559--1569, 2025{\natexlab{e}}.

\bibitem[O’Neill et~al.(2024)O’Neill, Rehman, Maddukuri, Gupta, Padalkar, Lee, Pooley, Gupta, Mandlekar, Jain, et~al.]{o2024open}
Abby O’Neill, Abdul Rehman, Abhiram Maddukuri, Abhishek Gupta, Abhishek Padalkar, Abraham Lee, Acorn Pooley, Agrim Gupta, Ajay Mandlekar, Ajinkya Jain, et~al.
\newblock Open x-embodiment: Robotic learning datasets and rt-x models: Open x-embodiment collaboration 0.
\newblock In \emph{ICRA}, 2024.

\bibitem[Pai et~al.(2025)Pai, Achenbach, Montesinos, Forrai, Mees, and Nava]{mimicvideo}
Jonas Pai, Liam Achenbach, Victoriano Montesinos, Benedek Forrai, Oier Mees, and Elvis Nava.
\newblock mimic-video: Video-action models for generalizable robot control beyond vlas.
\newblock \emph{arXiv preprint arXiv:2512.15692}, 2025.

\bibitem[Podell et~al.(2023)Podell, English, Lacey, Blattmann, Dockhorn, M{\"u}ller, Penna, and Rombach]{sdxl}
Dustin Podell, Zion English, Kyle Lacey, Andreas Blattmann, Tim Dockhorn, Jonas M{\"u}ller, Joe Penna, and Robin Rombach.
\newblock Sdxl: Improving latent diffusion models for high-resolution image synthesis.
\newblock \emph{arXiv preprint arXiv:2307.01952}, 2023.

\bibitem[Shen et~al.(2025)Shen, Wei, Du, Liang, Lu, Yang, Zheng, and Guo]{shen2025videovla}
Yichao Shen, Fangyun Wei, Zhiying Du, Yaobo Liang, Yan Lu, Jiaolong Yang, Nanning Zheng, and Baining Guo.
\newblock Videovla: Video generators can be generalizable robot manipulators.
\newblock \emph{arXiv preprint arXiv:2512.06963}, 2025.

\bibitem[Steiner et~al.(2024)Steiner, Pinto, Tschannen, Keysers, Wang, Bitton, Gritsenko, Minderer, Sherbondy, Long, et~al.]{paligemma2}
Andreas Steiner, Andr{\'e}~Susano Pinto, Michael Tschannen, Daniel Keysers, Xiao Wang, Yonatan Bitton, Alexey Gritsenko, Matthias Minderer, Anthony Sherbondy, Shangbang Long, et~al.
\newblock Paligemma 2: A family of versatile vlms for transfer.
\newblock \emph{arXiv preprint arXiv:2412.03555}, 2024.

\bibitem[Team et~al.(2025{\natexlab{a}})Team, Zhu, Wang, Zhou, Chang, Zhou, Li, Chen, Shen, Pang, et~al.]{team2025aether}
Aether Team, Haoyi Zhu, Yifan Wang, Jianjun Zhou, Wenzheng Chang, Yang Zhou, Zizun Li, Junyi Chen, Chunhua Shen, Jiangmiao Pang, et~al.
\newblock Aether: Geometric-aware unified world modeling.
\newblock \emph{arXiv preprint arXiv:2503.18945}, 2025{\natexlab{a}}.

\bibitem[Team et~al.(2025{\natexlab{b}})Team, Ye, Wang, Ni, Huang, Zhao, Li, Li, Zhu, Feng, et~al.]{gigabrain0}
GigaBrain Team, Angen Ye, Boyuan Wang, Chaojun Ni, Guan Huang, Guosheng Zhao, Haoyun Li, Jie Li, Jiagang Zhu, Lv~Feng, et~al.
\newblock Gigabrain-0: A world model-powered vision-language-action model.
\newblock \emph{arXiv preprint arXiv:2510.19430}, 2025{\natexlab{b}}.

\bibitem[Team et~al.(2026)Team, Wang, Ni, Huang, Zhao, Li, Li, Lv, Liu, Feng, et~al.]{team2026gigabrain05}
GigaBrain Team, Boyuan Wang, Chaojun Ni, Guan Huang, Guosheng Zhao, Hao Li, Jie Li, Jindi Lv, Jingyu Liu, Lv~Feng, et~al.
\newblock Gigabrain-0.5 m*: a vla that learns from world model-based reinforcement learning.
\newblock \emph{arXiv preprint arXiv:2602.12099}, 2026.

\bibitem[Team et~al.(2025{\natexlab{c}})Team, Ye, Wang, Ni, Huang, Zhao, Li, Zhu, Li, Xu, et~al.]{gigaworld0}
GigaWorld Team, Angen Ye, Boyuan Wang, Chaojun Ni, Guan Huang, Guosheng Zhao, Haoyun Li, Jiagang Zhu, Kerui Li, Mengyuan Xu, et~al.
\newblock Gigaworld-0: World models as data engine to empower embodied ai.
\newblock \emph{arXiv preprint arXiv:2511.19861}, 2025{\natexlab{c}}.

\bibitem[Tian et~al.(2025)Tian, Wang, Zhou, Wang, Li, Sun, and Tang]{tian2025pdfactor}
Jingyi Tian, Le~Wang, Sanping Zhou, Sen Wang, Jiayi Li, Haowen Sun, and Wei Tang.
\newblock Pdfactor: Learning tri-perspective view policy diffusion field for multi-task robotic manipulation.
\newblock In \emph{Proceedings of the Computer Vision and Pattern Recognition Conference}, pages 15757--15767, 2025.

\bibitem[Wan et~al.(2025)Wan, Wang, Ai, Wen, Mao, Xie, Chen, Yu, Zhao, Yang, Zeng, Wang, Zhang, Zhou, Wang, Chen, Zhu, Zhao, Yan, Huang, Feng, Zhang, Li, Wu, Chu, Feng, Zhang, Sun, Fang, Wang, Gui, Weng, Shen, Lin, Wang, Wang, Zhou, Wang, Shen, Yu, Shi, Huang, Xu, Kou, Lv, Li, Liu, Wang, Zhang, Huang, Li, Wu, Liu, Pan, Zheng, Hong, Shi, Feng, Jiang, Han, Wu, and Liu]{wan}
Team Wan, Ang Wang, Baole Ai, Bin Wen, Chaojie Mao, Chen-Wei Xie, Di~Chen, Feiwu Yu, Haiming Zhao, Jianxiao Yang, Jianyuan Zeng, Jiayu Wang, Jingfeng Zhang, Jingren Zhou, Jinkai Wang, Jixuan Chen, Kai Zhu, Kang Zhao, Keyu Yan, Lianghua Huang, Mengyang Feng, Ningyi Zhang, Pandeng Li, Pingyu Wu, Ruihang Chu, Ruili Feng, Shiwei Zhang, Siyang Sun, Tao Fang, Tianxing Wang, Tianyi Gui, Tingyu Weng, Tong Shen, Wei Lin, Wei Wang, Wei Wang, Wenmeng Zhou, Wente Wang, Wenting Shen, Wenyuan Yu, Xianzhong Shi, Xiaoming Huang, Xin Xu, Yan Kou, Yangyu Lv, Yifei Li, Yijing Liu, Yiming Wang, Yingya Zhang, Yitong Huang, Yong Li, You Wu, Yu~Liu, Yulin Pan, Yun Zheng, Yuntao Hong, Yupeng Shi, Yutong Feng, Zeyinzi Jiang, Zhen Han, Zhi-Fan Wu, and Ziyu Liu.
\newblock Wan: Open and advanced large-scale video generative models.
\newblock \emph{arXiv preprint arXiv:2503.20314}, 2025.

\bibitem[Wang et~al.(2025{\natexlab{a}})Wang, Meng, Wang, Zhu, Ye, Wang, Yang, Ni, Huang, and Wang]{embodiedreamer}
Boyuan Wang, Xinpan Meng, Xiaofeng Wang, Zheng Zhu, Angen Ye, Yang Wang, Zhiqin Yang, Chaojun Ni, Guan Huang, and Xingang Wang.
\newblock Embodiedreamer: Advancing real2sim2real transfer for policy training via embodied world modeling.
\newblock \emph{arXiv preprint arXiv:2507.05198}, 2025{\natexlab{a}}.

\bibitem[Wang et~al.(2025{\natexlab{b}})Wang, Ouyang, Wang, Zhu, Zhao, Ni, Huang, Liu, and Wang]{humandreamerx}
Boyuan Wang, Runqi Ouyang, Xiaofeng Wang, Zheng Zhu, Guosheng Zhao, Chaojun Ni, Guan Huang, Lihong Liu, and Xingang Wang.
\newblock Humandreamer-x: Photorealistic single-image human avatars reconstruction via gaussian restoration.
\newblock \emph{arXiv preprint arXiv:2504.03536}, 2025{\natexlab{b}}.

\bibitem[Wang et~al.(2025{\natexlab{c}})Wang, Wang, Ni, Zhao, Yang, Zhu, Zhang, Zhou, Chen, Huang, Liu, and Wang]{humandreamer}
Boyuan Wang, Xiaofeng Wang, Chaojun Ni, Guosheng Zhao, Zhiqin Yang, Zheng Zhu, Muyang Zhang, Yukun Zhou, Xinze Chen, Guan Huang, Lihong Liu, and Xingang Wang.
\newblock Humandreamer: Generating controllable human-motion videos via decoupled generation.
\newblock \emph{arXiv preprint arXiv:2503.24026}, 2025{\natexlab{c}}.

\bibitem[Wang et~al.(2025{\natexlab{d}})Wang, Tian, Wang, Liao, Li, Dong, Xia, Zhou, Tang, and Gang]{wang2025sampo}
Sen Wang, Jingyi Tian, Le~Wang, Zhimin Liao, Jiayi Li, Huaiyi Dong, Kun Xia, Sanping Zhou, Wei Tang, and Hua Gang.
\newblock Sampo: Scale-wise autoregression with motion prompt for generative world models.
\newblock \emph{arXiv preprint arXiv:2509.15536}, 2025{\natexlab{d}}.

\bibitem[Wang et~al.(2025{\natexlab{e}})Wang, Wang, Zhou, Tian, Li, Sun, and Tang]{wang2025flowram}
Sen Wang, Le~Wang, Sanping Zhou, Jingyi Tian, Jiayi Li, Haowen Sun, and Wei Tang.
\newblock Flowram: Grounding flow matching policy with region-aware mamba framework for robotic manipulation.
\newblock In \emph{Proceedings of the Computer Vision and Pattern Recognition Conference}, pages 12176--12186, 2025{\natexlab{e}}.

\bibitem[Wang et~al.(2025{\natexlab{f}})Wang, Zhu, Zhang, Wang, Zhu, Zhao, Ni, Wang, Huang, Chen, et~al.]{wang2025drivegen3d}
Weijie Wang, Jiagang Zhu, Zeyu Zhang, Xiaofeng Wang, Zheng Zhu, Guosheng Zhao, Chaojun Ni, Haoxiao Wang, Guan Huang, Xinze Chen, et~al.
\newblock Drivegen3d: Boosting feed-forward driving scene generation with efficient video diffusion.
\newblock \emph{arXiv preprint arXiv:2510.15264}, 2025{\natexlab{f}}.

\bibitem[Wang et~al.(2025{\natexlab{g}})Wang, Ding, Li, Cui, Ge, Tong, Song, Zhao, Zhao, Hou, Huang, Tang, Wang, Zhang, Liu, and Wang]{wang2025vlaadapter}
Yihao Wang, Pengxiang Ding, Lingxiao Li, Can Cui, Zirui Ge, Xinyang Tong, Wenxuan Song, Han Zhao, Wei Zhao, Pengxu Hou, Siteng Huang, Yifan Tang, Wenhui Wang, Ru~Zhang, Jianyi Liu, and Donglin Wang.
\newblock Vla-adapter: An effective paradigm for tiny-scale vision-language-action model.
\newblock \emph{arXiv preprint arXiv:2509.09372}, 2025{\natexlab{g}}.

\bibitem[Wu et~al.(2025)Wu, Li, and Su]{11357220}
Hao Wu, Hui Li, and Yiyun Su.
\newblock Bridging the perception-cognition gap:re-engineering sam2 with hilbert-mamba for robust vlm-based medical diagnosis.
\newblock In \emph{2025 IEEE International Conference on Bioinformatics and Biomedicine (BIBM)}, pages 4275--4278, 2025.

\bibitem[Wu et~al.(2024)Wu, Hou, Liu, Che, Ju, Yang, Li, Zhao, Xu, Yang, et~al.]{robomind}
Kun Wu, Chengkai Hou, Jiaming Liu, Zhengping Che, Xiaozhu Ju, Zhuqin Yang, Meng Li, Yinuo Zhao, Zhiyuan Xu, Guang Yang, et~al.
\newblock Robomind: Benchmark on multi-embodiment intelligence normative data for robot manipulation.
\newblock \emph{arXiv preprint arXiv:2412.13877}, 2024.

\bibitem[Xiang et~al.(2024)Xiang, Liu, Gu, Gao, Ning, Zha, Feng, Tao, Hao, Shi, et~al.]{xiang2024pandora}
Jiannan Xiang, Guangyi Liu, Yi~Gu, Qiyue Gao, Yuting Ning, Yuheng Zha, Zeyu Feng, Tianhua Tao, Shibo Hao, Yemin Shi, et~al.
\newblock Pandora: Towards general world model with natural language actions and video states.
\newblock \emph{arXiv preprint arXiv:2406.09455}, 2024.

\bibitem[Ye et~al.(2025)Ye, Zhang, Wang, Wang, Zhang, and Zhu]{vlar1}
Angen Ye, Zeyu Zhang, Boyuan Wang, Xiaofeng Wang, Dapeng Zhang, and Zheng Zhu.
\newblock Vla-r1: Enhancing reasoning in vision-language-action models.
\newblock \emph{arXiv preprint arXiv:2510.01623}, 2025.

\bibitem[Ye et~al.(2026{\natexlab{a}})Ye, Chen, Zhong, Xiao, Zhang, Wu, and Shen]{SEExiao}
Hua Ye, Siyuan Chen, Ziqi Zhong, Canran Xiao, Haoliang Zhang, Yuhan Wu, and Fei Shen.
\newblock Seeing through the conflict: Transparent knowledge conflict handling in retrieval-augmented generation.
\newblock \emph{arXiv preprint arXiv:2601.06842}, 2026{\natexlab{a}}.

\bibitem[Ye et~al.(2026{\natexlab{b}})Ye, Ge, Zheng, Gao, Yu, Kurian, Indupuru, Tan, Zhu, Xiang, Malik, Lee, Liang, Ranawaka, Gu, Xu, Wang, Hu, Narayan, Bjorck, Wang, Kim, Niu, Zheng, Xie, Wu, Wang, Julian, Xu, Du, Chebotar, Reed, Kautz, Zhu, Fan, and Jang]{ye2026worldactionmodelszeroshot}
Seonghyeon Ye, Yunhao Ge, Kaiyuan Zheng, Shenyuan Gao, Sihyun Yu, George Kurian, Suneel Indupuru, You~Liang Tan, Chuning Zhu, Jiannan Xiang, Ayaan Malik, Kyungmin Lee, William Liang, Nadun Ranawaka, Jiasheng Gu, Yinzhen Xu, Guanzhi Wang, Fengyuan Hu, Avnish Narayan, Johan Bjorck, Jing Wang, Gwanghyun Kim, Dantong Niu, Ruijie Zheng, Yuqi Xie, Jimmy Wu, Qi~Wang, Ryan Julian, Danfei Xu, Yilun Du, Yevgen Chebotar, Scott Reed, Jan Kautz, Yuke Zhu, Linxi~"Jim" Fan, and Joel Jang.
\newblock World action models are zero-shot policies, 2026{\natexlab{b}}.
\newblock URL \url{https://arxiv.org/abs/2602.15922}.

\bibitem[Zhang et~al.(2025)Zhang, Liu, Qi, Wang, Yu, Zhang, Dong, He, Wang, Zhang, et~al.]{dreamvla}
Wenyao Zhang, Hongsi Liu, Zekun Qi, Yunnan Wang, Xinqiang Yu, Jiazhao Zhang, Runpei Dong, Jiawei He, He~Wang, Zhizheng Zhang, et~al.
\newblock Dreamvla: a vision-language-action model dreamed with comprehensive world knowledge.
\newblock \emph{arXiv preprint arXiv:2507.04447}, 2025.

\bibitem[Zhao et~al.(2025{\natexlab{a}})Zhao, Ni, Wang, Zhu, Zhang, Wang, Huang, Chen, Wang, Zhang, et~al.]{drivedreamer4d}
Guosheng Zhao, Chaojun Ni, Xiaofeng Wang, Zheng Zhu, Xueyang Zhang, Yida Wang, Guan Huang, Xinze Chen, Boyuan Wang, Youyi Zhang, et~al.
\newblock Drivedreamer4d: World models are effective data machines for 4d driving scene representation.
\newblock In \emph{Proceedings of the computer vision and pattern recognition conference}, pages 12015--12026, 2025{\natexlab{a}}.

\bibitem[Zhao et~al.(2025{\natexlab{b}})Zhao, Wang, Ni, Zhu, Qin, Huang, and Wang]{recondreamer++}
Guosheng Zhao, Xiaofeng Wang, Chaojun Ni, Zheng Zhu, Wenkang Qin, Guan Huang, and Xingang Wang.
\newblock Recondreamer++: Harmonizing generative and reconstructive models for driving scene representation.
\newblock \emph{arXiv preprint arXiv:2503.18438}, 2025{\natexlab{b}}.

\bibitem[Zheng et~al.(2025)Zheng, Li, Wang, Liu, Kang, Feng, Zheng, Zou, Chen, Zeng, et~al.]{zheng2025x}
Jinliang Zheng, Jianxiong Li, Zhihao Wang, Dongxiu Liu, Xirui Kang, Yuchun Feng, Yinan Zheng, Jiayin Zou, Yilun Chen, Jia Zeng, et~al.
\newblock X-vla: Soft-prompted transformer as scalable cross-embodiment vision-language-action model.
\newblock \emph{arXiv preprint arXiv:2510.10274}, 2025.

\bibitem[Zhou et~al.(2024)Zhou, Du, Chen, Li, Yeung, and Gan]{robodreamer}
Siyuan Zhou, Yilun Du, Jiaben Chen, Yandong Li, Dit-Yan Yeung, and Chuang Gan.
\newblock Robodreamer: Learning compositional world models for robot imagination.
\newblock \emph{arXiv preprint arXiv:2404.12377}, 2024.

\end{thebibliography}

\end{document}